\documentclass[10pt,journal,compsoc]{IEEEtran}

\ifCLASSOPTIONcompsoc
  \usepackage[nocompress]{cite}
\else
  \usepackage{cite}
\fi

\usepackage{graphicx}
\usepackage{amsmath,amssymb}
\usepackage[linesnumbered,ruled]{algorithm2e}
\usepackage{multirow}
\usepackage{color}
\usepackage{spverbatim}
\usepackage{etoolbox}
\makeatletter
\preto{\@verbatim}{\topsep=1pt \partopsep=1pt}
\makeatother

\usepackage{array}
\newcolumntype{L}[1]{>{\raggedright\let\newline\\\arraybackslash\hspace{0pt}}m{#1}}
\newcolumntype{C}[1]{>{\centering\let\newline\\\arraybackslash\hspace{0pt}}m{#1}}
\newcolumntype{R}[1]{>{\raggedleft\let\newline\\\arraybackslash\hspace{0pt}}m{#1}}

\begin{document}

\title{Weight-Sharing Neural Architecture Search:\\A Battle to Shrink the Optimization Gap}

\author{Lingxi~Xie,
        Xin~Chen,
        Kaifeng~Bi,
        Longhui~Wei,
        Yuhui~Xu,
        Zhengsu~Chen,
        Lanfei~Wang,
        An~Xiao,
        Jianlong~Chang,
        Xiaopeng~Zhang,
        and~Qi~Tian,~\IEEEmembership{Fellow,~IEEE}
\IEEEcompsocitemizethanks{
\IEEEcompsocthanksitem Lingxi Xie, Xin Chen, Kaifeng Bi, Longhui Wei, Zhengsu Chen, An Xiao, Jianlong Chang, Jianlong Chang, Xiaopeng Zhang, and Qi Tian are with Huawei Inc., China.\protect\\
E-mail: {198808xc@gmail.com}, {1410452@tongji.edu.cn}, {bikaifeng1@huawei.com}, {weilonghui1@huawei.com}, {chenzhengsu1@huawei.com}, {xiaoan1@huawei.com}, {jianlong.chang@huawei.com}, {zxphistory@gmail.com}, {tian.qi1@huawei.com}
\IEEEcompsocthanksitem Yuhui Xu is with Shanghai Jiao Tong University, China.\protect\\
E-mail: yuhuixu@sjtu.edu.cn
\IEEEcompsocthanksitem Lanfei Wang is with Beijing University of Posts and Telecommunications, China.\protect\\
E-mail: wanglanfei@bupt.edu.cn}
\thanks{Manuscript received Month Date, 2020.}}

\markboth{Manuscript Draft: July 31st, 2020}
{Xie \MakeLowercase{\textit{et al.}}: Weight-Sharing Neural Architecture Search: A Battle to Shrink the Optimization Gap}

\IEEEtitleabstractindextext{%
\begin{abstract}
Neural architecture search (NAS) has attracted increasing attentions in both academia and industry. In the early age, researchers mostly applied \textbf{individual} search methods which sample and evaluate the candidate architectures separately and thus incur heavy computational overheads. To alleviate the burden, \textbf{weight-sharing} methods were proposed in which exponentially many architectures share weights in the same \textbf{super-network}, and the costly training procedure is performed only once. These methods, though being much faster, often suffer the issue of instability. This paper provides a literature review on NAS, in particular the weight-sharing methods, and points out that the major challenge comes from the \textbf{optimization gap} between the super-network and the sub-architectures. From this perspective, we summarize existing approaches into several categories according to their efforts in bridging the gap, and analyze both advantages and disadvantages of these methodologies. Finally, we share our opinions on the future directions of NAS and AutoML. Due to the expertise of the authors, this paper mainly focuses on the application of NAS to computer vision problems and may bias towards the work in our group.
\end{abstract}

}

\maketitle

\IEEEdisplaynontitleabstractindextext

\IEEEpeerreviewmaketitle

\IEEEraisesectionheading{
\section{Introduction}
\label{introduction}}

\IEEEPARstart{T}{he} rapid development of deep learning~\cite{lecun2015deep} has claimed its domination in the area of artificial intelligence. In particular, in the computer vision community, deep neural networks have been successfully applied to a wide range of challenging problems including image classification~\cite{krizhevsky2012imagenet,simonyan2015very,szegedy2015going,he2016deep,huang2017densely}, object detection~\cite{girshick2015fast,ren2015faster,redmon2016you,liu2016ssd}, semantic segmentation~\cite{long2015fully,chen2017deeplab,he2017mask}, boundary detection~\cite{xie2015holistically}, pose estimation~\cite{toshev2014deeppose}, \textit{etc}., and surpassed conventional approaches based on hand-designed features (\textit{e.g.}, SIFT~\cite{lowe2004distinctive}) by significant margins.

Despite the success of these efforts, researchers quickly encountered the difficulty that the network design space has become much more complex than that in the early deep learning era. Several troubles emerged. For example, the manually designed networks do not to generalize to various scenarios, \textit{e.g.}, a network designed for holistic image classification may produce inferior performance in dense image prediction tasks (semantic segmentation, image super-resolution, \textit{etc.}). In addition, it is sometimes required to design networks under specific hardware constraints (\textit{e.g.}, FLOPs, memory, latency, \textit{etc.}), and manual design is often insufficient to explore a large number of possibilities. Motivated by these factors, researchers started to seek for the solution that can flexibly adjusted to different scenarios in minimal efforts.

\textbf{Automated machine learning} (AutoML~\cite{thornton2013auto}) appeared recently as a practical tool which enables researchers and even engineers without expertise in machine learning to use off-the-shelf algorithms conveniently for different purposes. As a subarea of AutoML, \textbf{neural architecture search} (NAS) focuses on designing effective neural networks in an automatic manner~\cite{real2017large,xie2017genetic,zoph2017neural}. By defining a large set of possible network architectures (the set is often referred to as the search space), NAS can explore and evaluate a large number of networks which have never been studied before. Nowadays, NAS has achieved remarkable performance gain beyond manually designed architectures, in particular, in the scenarios that computational costs are constrained, \textit{e.g.}, in the \textsf{mobile setting}~\cite{howard2017mobilenets} of ImageNet classification.

In the early age, NAS algorithms often sampled a large number of architectures from the search space and then trained each of them from scratch to validate its performance. Despite their effectiveness in finding high-quality architectures, these approaches often require heavy computational overhead (\textit{e.g.}, tens of thousands of GPU days~\cite{zoph2017neural}), which hinders them from being transplanted to various applications. To alleviate this issue, researchers proposed to re-use the weights of previously optimized architectures~\cite{cai2018efficient} or share computation among different but relevant architectures~\cite{pham2018efficient} sampled from the search space. This direction eventually led to the idea that the search space is formulated into an over-parameterized \textbf{super-network} so that the sampled \textbf{sub-architectures} get evaluated without additional optimization~\cite{brock2017smash,liu2019darts}. Consequently, the search costs have been reduced by several orders of magnitudes, \textit{e.g.}, recent advances allow an effective architecture to be found within $0.1$ GPU-days on the CIFAR10 dataset~\cite{xu2020pc}, or within $2.0$ GPU-days on the ImageNet dataset~\cite{bi2020gold}.

This paper focuses on the weight-sharing methodology for NAS, and investigate the critical weakness of it, \textbf{instability}, which refers to the phenomenon that individual runs of search can lead to very different performance. We owe the instability to the \textbf{optimization gap} between the super-network and its sub-architectures -- in other words, the search procedure tries to optimize the super-network, but a well-optimized super-network does not necessarily produce high-quality sub-architectures. This is somewhat similar to the well-known over-fitting problem in machine learning, and thus can be alleviated from different aspects. We try to summarize the existing efforts into a unified framework, and thus we can answer the questions of where the community is, what the main difficulties are, what to do in the near future.

We noticed previous surveys~\cite{elsken2019neural,he2019automl,jaafra2019reinforcement,wistuba2019survey,ren2020comprehensive,talbi2020optimization} which used three factors (\textit{i.e.}, search space, search strategy, evaluation method) to categorize NAS approaches. We inherit this framework but merges the latter two factors since they are highly coupled in recent works. Differently, this paper focuses on the weight-sharing methodology and investigate it from the perspective of shrinking the optimization gap. As a disclaimer, this paper discusses the modern neural architecture search methods, \textit{i.e.}, in the context of deep learning, so we do not cover the early work on finding efficient neural networks~\cite{yao1999evolving,stanley2002evolving,bayer2009evolving,ding2013evolutionary}.

The remainder of this paper is organized as follows. Section~\ref{formulation} first formulates NAS into the problem of finding the optimal sub-architecture(s) in a super-network, and introduces both individual and weight-sharing approaches to explore the search space efficiently. Then, Section~\ref{weight_sharing} is the main part of this paper, in which we formalize the optimization gap to be the main challenge of weight-sharing NAS, based on which we review a few popular but preliminary solutions to shrink the gap. Next, in Section~\ref{others}, we discuss a few other directions that are related to NAS. Finally, in Section~\ref{future}, we look into the future of NAS and AutoML, and put forward a few open problems in the research field.

\section{Neural Architecture Search}
\label{formulation}

In this section, we provide a generalized formulation for neural architecture search (NAS) and build up a notation system which will be used throughout this paper. We first introduce the fundamental concepts of NAS in Section~\ref{formulation:framework}. Then, two basic elements, the search space and the search strategy (including the evaluation method), are introduced in Sections~\ref{formulation:space} and~\ref{formulation:search_evaluation}. Lastly, there is a brief summary in Section~\ref{formulation:summary}.

\subsection{Overall Framework}
\label{formulation:framework}

The background of NAS is identical to regular machine learning tasks, in which a dataset $\mathcal{D}$ with $N$ elements is given, and each element in it is denoted by an input-output pair, $\left(\mathbf{x}_n,\mathbf{y}_n\right)$. The goal is to find a model that works best on the dataset and, more importantly, transfers well to the testing scenario. For later convenience, we use $\mathcal{D}_\mathrm{train}$, $\mathcal{D}_\mathrm{val}$ and $\mathcal{D}_\mathrm{test}$ to indicate the training, validation and testing subsets of $\mathcal{D}$, respectively. There are some standard datasets for NAS. For image classification, the most popular datasets are CIFAR10~\cite{krizhevsky2009learning} and ImageNet-1K~\cite{deng2009imagenet,russakovsky2015imagenet}. For natural language modeling, the most popular ones are Penn Treebank and WikiText-2~\cite{merity2016pointer}.

NAS starts with defining a \textbf{search space}, $\mathcal{S}$. Each element ${\mathbb{S}}\in{\mathcal{S}}$ is a network architecture that receives an input $\mathbf{x}$ and delivers an output $\mathbf{y}$, which we denote as ${\mathbf{y}}={\mathbf{f}^\mathbb{S}\!\left(\mathbf{x};\boldsymbol{\omega}\right)}$ where ${\boldsymbol{\omega}}\doteq{\boldsymbol{\omega}\!\left(\mathbb{S}\right)}$ indicates the network weights (\textit{e.g.}, in the convolutional layers). Note that the form and dimensionality of $\boldsymbol{\omega}$ depends on $\mathbb{S}$. We use $\left\langle\mathbb{S},\boldsymbol{\omega}^\star\!\left(\mathbb{S}\right)\right\rangle$ to denote an architecture equipped with the optimal network weights. The goal of NAS is to find the optimal architecture $\mathbb{S}^\star$ which produces the best performance (\textit{e.g.}, classification accuracy) on the testing set:
\begin{equation}
\label{eqn:goal}
{\mathbb{S}^\star}={\arg\max_{\mathbb{S}\in\mathcal{S}}\mathrm{Eval}\!\left(\left\langle\mathbb{S},\boldsymbol{\omega}^\star\!\left(\mathbb{S}\right)\right\rangle;\mathcal{D}_\mathrm{test}\right)},
\end{equation}
where $\mathrm{Eval}\!\left(\left\langle\mathbb{S},\boldsymbol{\omega}^\star\!\left(\mathbb{S}\right)\right\rangle\middle\vert\mathcal{D}_\mathrm{test}\right)$ is the function that measures how $\left\langle\mathbb{S},\boldsymbol{\omega}^\star\!\left(\mathbb{S}\right)\right\rangle$ behaves on the testing set, $\mathcal{D}_\mathrm{test}$. But, in real practice, $\mathcal{D}_\mathrm{test}$ is hidden in the search procedure, so the training and validation sets are often used instead, and thus the objective becomes:
\begin{eqnarray}
\label{eqn:goal_trainval}
{} & {\mathbb{S}^\star}={\arg\max_{\mathbb{S}\in\mathcal{S}}\mathrm{Eval}\!\left(\left\langle\mathbb{S},\boldsymbol{\omega}^\star\!\left(\mathbb{S}\right)\right\rangle;\mathcal{D}_\mathrm{val}\right)}\\
\nonumber
\mathrm{s.t.} & {\boldsymbol{\omega}^\star\!\left(\mathbb{S}\right)}={\arg\min_{\boldsymbol{\omega}}\mathcal{L}\!\left(\boldsymbol{\omega}\!\left(\mathbb{S}\right);\mathcal{D}_\mathrm{train}\right)},
\end{eqnarray}
where $\mathcal{L}\!\left(\boldsymbol{\omega}\!\left(\mathbb{S}\right);\mathcal{D}_\mathrm{train}\right)$ is the loss function (\textit{e.g.}, cross-entropy) computed on the training set.

Most often, the search space $\mathcal{S}$ is sufficiently large to enable powerful architectures to be found. Therefore, it is unlikely that an exhaustive search method can explore the large space with reasonable computational costs. Therefore, a practical solution is to use a heuristic search method which defines $\mathfrak{P}$, a probabilistic distribution on $\mathcal{S}$, and adjusts the probability, $p\!\left(\mathbb{S}\right)$, that an element $\mathbb{S}$ is sampled according to the rewards from those already sampled architectures. This is to say, there is a more important factor, the \textbf{search strategy}, which determines the way of adjusting the probabilistic distribution. Note that each search strategy is equipped with an \textbf{evaluation method} which judges the quality of a sampled architecture.

A generalized framework of NAS is described in Algorithm~\ref{alg:pipeline}. In what follows, we briefly review the popular choices of the search space and search strategy.

\begin{algorithm}[!t]
\SetKwInOut{Input}{Input}
\SetKwInOut{Output}{Output}
\SetKwInOut{Return}{Return}
\Input{
The search space $\mathcal{S}$, the training and validation datasets $\mathcal{D}_\mathrm{train}$ and $\mathcal{D}_\mathrm{val}$, the evaluation method $\mathrm{Eval}\!\left(\left\langle\mathbb{S},\boldsymbol{\omega}\!\left(\mathbb{S}\right)\right\rangle;\mathcal{D}_\mathrm{val}\right)$;
}
\Output{
The optimal model $\left\langle\mathbb{S}^\star,\boldsymbol{\omega}^\star\!\left(\mathbb{S}^\star\right)\right\rangle$;
}
Initialize a probabilistic distribution $\mathfrak{P}$ on $\mathcal{S}$;\\
\Repeat{convergence \textbf{or} the time limit is reached}{
Sample an architecture $\mathbb{S}$ from $\mathfrak{P}$;\\
Train $\mathbb{S}$ on $\mathcal{D}_\mathrm{train}$ to obtain $\left\langle\mathbb{S},\boldsymbol{\omega}^\star\!\left(\mathbb{S}\right)\right\rangle$;\\
Compute ${\mathrm{Acc}}={\mathrm{Eval}\!\left(\left\langle\mathbb{S},\boldsymbol{\omega}^\star\!\left(\mathbb{S}\right)\right\rangle;\mathcal{D}_\mathrm{val}\right)}$;\\
\If{$\mathrm{Acc}$ surpasses the best seen value}{
Update the best model as $\left\langle\mathbb{S},\boldsymbol{\omega}^\star\!\left(\mathbb{S}\right)\right\rangle$;
}
Using the evaluation result to update $\mathfrak{P}$;\\
}
\Return{
The best seen model $\left\langle\mathbb{S}^\star,\boldsymbol{\omega}^\star\!\left(\mathbb{S}^\star\right)\right\rangle$.
}
\caption{A Generalized Pipeline of NAS}
\label{alg:pipeline}
\end{algorithm}

\begin{figure*}
\centering
\includegraphics[width=16cm]{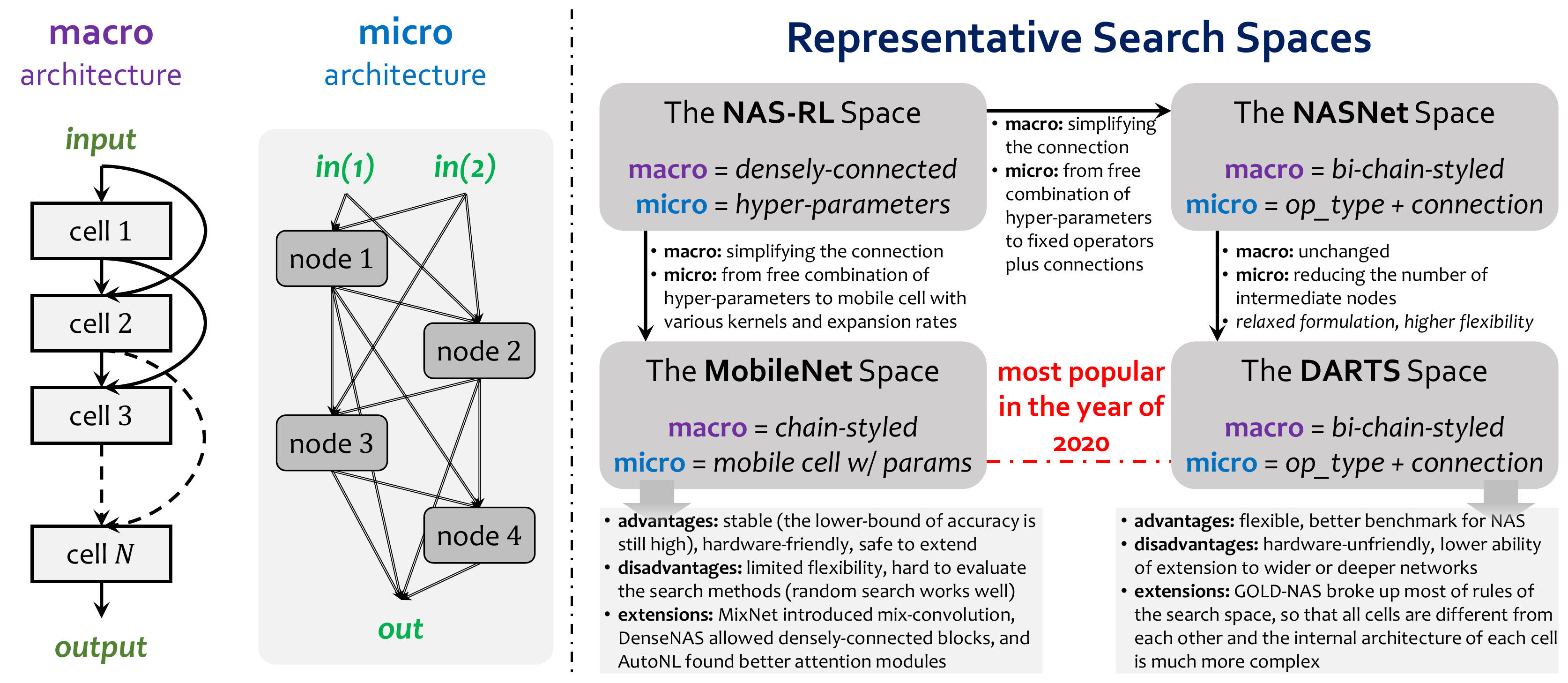}
\caption{\textbf{Left}: an example of the search space. For simplicity, we show a macro architecture where each cell is allowed to receive input from two precursors, and a micro architecture (cell) where the type and existence of the inter-node connections are searchable. \textbf{Right}: the relationship between four representative search spaces introduced in Sections~\ref{formulation:space:NASRL}--\ref{formulation:space:MobileNet}, respectively. For the two most popular search spaces to date, MobileNet and DARTS, we summarize the advantages and disadvantages below, and point out some extensions.}
\label{fig:search_spaces}
\end{figure*}

\subsection{Search Space}
\label{formulation:space}

The search space, $\mathcal{S}$, determines the range of architecture search. Increasing the scale of the search space, $\left|\mathcal{S}\right|$, is both intriguing and challenging. On the one hand, a larger space enables the search algorithm to cover more architectures so that the chance of finding a powerful architecture is increased. On the other hand, a larger search space makes the search algorithm more difficult to converge, because $\mathrm{Eval}\!\left(\left\langle\mathbb{S},\boldsymbol{\omega}\right\rangle;\mathcal{D}_\mathrm{val}\right)$ can be a very complicated function with respect to $\mathbb{S}$ -- more importantly, there is no guarantee of any of its properties in optimization. Hence, the number of samples required for depicting the function is often positively correlated to $\left|\mathcal{S}\right|$.

There exist many types of search spaces. There are often two steps of constructing a search space, namely, the macro and micro architectures of the space. As shown in Figure~\ref{fig:search_spaces}, the \textbf{macro} architecture determines the overall backbone of the network, and the \textbf{micro} architecture determines the details of each network unit. Sometimes, the micro architectures are referred to as \textbf{cells} and the number of cells can vary between the search and re-training phases or to adjust different tasks. Each NAS algorithm searches for either the macro or micro architecture but not necessarily both of them. Today, most NAS methods fix the macro architecture and investigate better micro architectures, which largely constrains the flexibility of the search space.

In what follows, we instantiate the macro and micro architectures using some popular examples. Note that some methods considered different search spaces for vision and language modeling and we will mainly focus on the vision part for which the search space is often more complicated. The relationship among these popular search spaces is illustrated in Figure~\ref{fig:search_spaces}.

\subsubsection{The \textbf{EvoNet} Search Space}
\label{formulation:space:EvoNet}

EvoNet~\cite{real2017large} is one of the first works that investigated an open search space, \textit{i.e.}, the architectures are not limited within the combination of a fixed number of components, and instead, the architecture is allowed to grow arbitrarily with requirement. Thus, the search space is latently defined by a set of \textbf{actions} on the architecture, \textit{e.g.}, inserting a convolutional layer to a specific position, altering the kernel size of a convolutional layer, adding a skip connection between two layers, \textit{etc.}, and some actions can be reversely performed, \textit{e.g.}, removing the convolutional layer from a specific position.

In an open search space, there is not a fixed number of architectures because the network shape (\textit{e.g.}, depth) is only constrained by the hardware platform -- even provided the length of the search procedure and the hardware limit, the accurate space size is difficult to estimate. A very loose estimation is obtained by assuming that the search procedure starts with a fixed architecture and has $T$ steps in total. Thus, if each step can take up to $C$ actions, the search space size satisfies ${\left|\mathcal{S}\right|}\leqslant{1+C+C^2+\ldots+C^T}$.

\noindent\textbf{Other open search spaces.}\quad There are other efforts that tried to explore an open space. For example, \cite{liu2018hierarchical} formulated the search procedure in a hierarchical way and allow the connection between some layers to be replaced by small modules named motifs; and~\cite{wang2019sample} learned an action space to generate complex architectures. As we shall explain in Section~\ref{future}, open-space search is likely to be the future trend of NAS~\cite{real2020automl}, but the community needs a robust search strategy that can efficiently explore the challenging spaces.

\subsubsection{The \textbf{NAS-RL} Search Space}
\label{formulation:space:NASRL}

NAS-RL~\cite{zoph2017neural} is the work that defined the modern formulation of NAS. It also sculptured the framework that uses the macro and micro architectures to determine the network. The \textbf{densely-connected} macro architecture has $N$ layers, each of which can be connected to any of its precursors. Besides the direct precursor, other connections can be present or absent, bringing a total of $2^{\left(N-1\right)\left(N-2\right)/2}$ possibilities. Each layer has a micro architecture involving the hyper-parameters of the filter height, filter width, stride height, stride width, and the number of filters. For simplicity, each hyper-parameter is chosen from a pre-defined set, \textit{e.g.}, the number of filters is one among $\left\{24,36,48,64\right\}$. Let $C$ be the number of combinations for each layer, then the total number of architectures is ${\left|\mathcal{S}\right|}={C^N\times2^{N\left(N-1\right)/2}}$. In the original paper, $C$ varies with different search options and the largest $C$ is $576$.

\noindent\textbf{The search space for language modeling.}\quad The goal is to construct a cell that connects the neighboring states of a recurrent network, in which there are three input variables and two output variables. Similarly, the recurrent controller is used to determine the architecture with a fixed number of intermediate nodes. For each node that is related to the input data and the hidden state, there is an operator (\textit{e.g.}, \textsf{add}, \textsf{elem-mult}, \textit{etc.}) and an activation function (\textit{e.g.}, \textsf{tanh}, \textsf{ReLU}, \textsf{sigmoid}, \textit{etc.}), and the controller also determines how to connect the memory states to the temporary variables inside the tree.

\noindent\textbf{Related search spaces.}\quad Genetic-CNN~\cite{xie2017genetic} shared the similar ideology of building the architecture in a macro-then-micro manner. The architecture is composed of a fixed number of stages and each stage has a fixed number of layers. Pooling layers are inserted between neighboring stages. Only the regular \textsf{3x3-convolution} is used, but the inter-layer connectivity in the same stage can be searched. The number of architectures in each stage is $2^{N\left(N-1\right)/2}$ where $N$ is the number of layers in the stage.

\subsubsection{The \textbf{NASNet} Search Space}
\label{formulation:space:NASNet}

NASNet~\cite{zoph2018learning} introduced the idea that the macro architecture contains repeatable cells, so that the search procedure is performed within a relatively shallow network but the final architecture can be much deeper. In addition, the densely-connected macro architecture is simplified into \textbf{bi-chain-styled}, \textit{i.e.}, each layer steadily receives input from two precursors. Regarding the micro architecture, each cell contains $N$ hidden nodes, indexed from $0$ to $N-1$, and the two input nodes are indexed as $-2$ and $-1$. Each hidden node is connected to two nodes with smaller indices, with the operator of each connection chosen from a pre-defined set of $C_1$ candidates (\textit{e.g.}, \textsf{dil-conv-3x3}, \textsf{sep-conv-5x5}, \textsf{identity}, \textsf{max-pool-3x3}, \textit{etc.}) and the summarization function chosen from a pre-defined set of $C_2$ candidates (\textit{e.g.}, \textsf{sum}, \textsf{concat}, \textsf{product}, \textit{etc.}). The outputs of all intermediate nodes are concatenated into the output of the cell. $C_1$ is $13$ in NASNet, but is often reduced to a smaller number in the follow-up methods because researchers noticed that there exist some weak operators that are rarely used (\textit{e.g.}, \textsf{max-pool-7x7} and \textsf{1x7-then-7x1-conv}), plus, shrinking the search space can reduce the cost of search. We denote $C_3$ as the number of different topologies in a cell, which is computed via ${2\choose2}\times{3\choose2}\times\ldots\times{N+1\choose2}$. The number of architectures in a normal cell or a reduction cell is thus $\left(C_1^2\times C_2\right)^N\times C_3$.

\noindent\textbf{Related search spaces.}\quad Many follow-up methods have used the same search space but simplified either the candidate set or the cell topology. For example, PNASNet~\cite{liu2018progressive} and AmoebaNet~\cite{real2019regularized} reduced $C_1$ to $8$, and DARTS~\cite{liu2019darts} reduced $N$ from $5$ to $4$. In addition, DARTS did not use individual summarization functions for different nodes and used \textsf{sum} as the default choice.

\subsubsection{The \textbf{DARTS} Search Space}
\label{formulation:space:DARTS}

DARTS~\cite{liu2019darts} has a very similar search space to NASNet and follows a weight-sharing search pipeline (see Section~\ref{formulation:search_evaluation}) so that a more flexible super-network is constructed. In the standard DARTS space, the set of candidate operators have been defined as \textsf{sep-conv-3x3}, \textsf{sep-conv-5x5}, \textsf{dil-conv-3x3}, \textsf{dil-conv-5x5}, \textsf{max-pool-3x3}, \textsf{avg-pool-3x3}, and \textsf{skip-connect}. A dummy \textsf{zero} operator is added to the search procedure but is not allowed to appear in the final architecture. To ease the formulation of differentiable architecture search, each layer receives input data from an arbitrary number of previous layers in the same cell though only two of them are allowed to survive. Provided that $4$ intermediate cells are present, the number of architectures for both the normal and reduction cells is ${{2\choose2}{3\choose2}{4\choose2}{5\choose2}\times7^8}\approx{1.0\times10^9}$, and the entire space has around $1.1\times10^{18}$ architectures.

\noindent\textbf{The search space for language modeling.}\quad Similar to NAS-RL, the DARTS space is also used for finding an improved recurrent cell in the language models. The input sources, the number of intermediate nodes, and the connectivity between these nodes are adjusted accordingly, but still, the outputs of all intermediate nodes are concatenated.

\noindent\textbf{Related search spaces.}\quad The super-network formulation of DARTS has offered opportunities to create some new search spaces. For example, one can shrink the search space by reducing the number of operators (\textit{e.g.}, only \textsf{sep-conv-3x3} and \textsf{skip-connect} used~\cite{zela2020understanding}), or expanding the search space by enabling multiple operators to be preserved in each edge~\cite{zhou2019bayesnas,cho2019one}, adding new operators~\cite{wang2019scalable}, allowing the cells of the same type to be individual from each other~\cite{bi2019stabilizing}, and relax the constraints that each layer receives input from two previous layers~\cite{bi2020gold}.

\begin{table*}[!t]
\centering
\begin{tabular}{|L{3.5cm}||C{3cm}|C{9cm}|}
\hline
\textbf{Search Space / Task} & \textbf{Important Work} & \textbf{Other Work} \\
\hline\hline
\textbf{EvoNet}~\cite{real2017large} \& open &
\cite{real2017large,liu2018hierarchical} &
\cite{real2020automl} \\
\hline
\textbf{NAS-RL}~\cite{zoph2017neural} &
\cite{zoph2017neural} &
\cite{zhou2018resource,bashivan2019teacher} \\
\hline
\textbf{NASNet}~\cite{zoph2018learning} &
\cite{zoph2018learning,zhong2018practical,liu2018progressive,pham2018efficient,real2019regularized,bender2018understanding,luo2018neural} &
\cite{liang2018evolutionary,chen2019renas,perez2018efficient,zhong2020blockqnn,zhang2018you,guo2019irlas,laube2019shufflenasnets,wang2019alphax,macko2019improving,wistuba2019inductive,chen2019automatic,adam2019understanding,akimoto2019adaptive,saltori2019regularized,pasunuru2019continual,liu2019memnet,zhou2019epnas,shi2019multi,zhang2020autoshrink,ardywibowo2020nads} \\
\hline
\textbf{DARTS}~\cite{liu2019darts} &
\cite{liu2019darts,xie2019snas,li2019random,dong2019searching,chen2019progressive,zhou2019bayesnas,nayman2019xnas,xu2020pc,zela2020understanding,liang2019darts+,guo2019nat,dong2019one} &
\cite{casale2019probabilistic,hundt2019sharpdarts,noy2020asap,weng2019automatic,zheng2019multinomial,hu2019efficient,yao2020efficient,chang2019data,zheng2019dynamic,wang2019sample,cho2019one,zhang2019efficient,yang2020cars,li2019stacnas,carlucci2019manas,luo2019understanding,bi2019stabilizing,lee2020efficient,chu2020fair,li2020sgas,dong2020automatic,green2019rapdarts,hong2019edas,xie2019exploiting,jin2019rc,vahdat2020unas,wang2019scalable,xu2020latency,zhou2020econas,hu2020dsnas,singh2020learning,chen2020stabilizing,he2020milenas,liu2020labels,niu2020disturbance,bulat2020bats,zhang2020adwpnas,li2020geometry,chu2020noisy,zheng2020rethinking,zhuo2020cp,zaidi2020neural,chen2020drnas,wang2020m,yu2020cyclic,zhao2020few,zhou2020theory,bi2020gold,guo2020breaking,hong2020dropnas,kaplan2020self,li2020neural3,tian2020discretization,wang2020mergenas,wang2020si,zhang2020one} \\
\hline
\textbf{MobileNet}~\cite{cai2019proxylessnas,wu2019fbnet} &
\cite{tan2019mnasnet,cai2019proxylessnas,wu2019fbnet,dai2019chamnet,stamoulis2019single,guo2019single,tan2019efficientnet,chu2019fairnas,tan2019mixconv,cai2020once} &
\cite{fang2019eat,wong2019attonets,xiong2019resource,han2019design,fang2020densely,shaw2019squeezenas,chu2020moga,chu2019scarletnas,yan2019hm,sun2019automatic,yu2020bignas,li2020improving,li2020block,kang2020towards,mei2020atomnas,chen2019efficient,chu2020mixpath,ding2020bnas,dai2020data,huang2020ponas,shen2020bs,you2020greedynas,wan2020fbnetv2,li2020neural,chen2020fitting,hu2020angle,guo2020powering,lu2020neural,phan2020binarizing,wang2020dc,xia2020hnas,yang2020hournas,bender2020can,chaudhuri2020fine,dai2020fbnetv3,li2020gp,liu2020memnas,lyna2020fast,trofimov2020multi,lu2020nsganetv2} \\
\hline
\textbf{Other Spaces} &
\cite{cortes2017adanet,baker2017designing,xie2017genetic,suganuma2017genetic,negrinho2017deeparchitect,baker2017accelerating,cai2018efficient,brock2017smash,kandasamy2018neural,elsken2019efficient,stanley2019designing,xie2019exploring} &
\cite{smithson2016neural,wei2017modularized,wang2018evolving,elsken2017simple,sun2018evolving,sun2018particle,wistuba2017finding,dutta2018effective,shin2018differentiable,kamath2018neural,cai2018path,dong2018dpp,jin2018efficient,hsu2018monas,istrate2019tapas,wistuba2018deep,rohekar2018constructing,prellberg2018lamarckian,lorenzo2018memetic,sun2019evolving,wang2018hybrid,cheng2018searching,zhang2019graph,lu2019nsga,sun2018automatically,fielding2018evolving,geifman2019deep,cheng2020instanas,wu2018mixed,jiang2019accuracy,dikov2019bayesian,savarese2019learning,wang2019hybrid,junior2019particle,radosavovic2019network,saikia2019autodispnet,ma2019deep,zhu2019eena,nguyen2020constrained,jiang2019neural,sun2019surrogate,cheng2019swiftnet,zhou2020posterior,wang2019particle,martens2019neural,irwin2019graph,shen2019searching,balaprakash2019scalable,zhao2019efficient,cheng2019msnet,hassantabar2019steerage,jiang2020efficient,meng2020neural,cheng2020nasgem} \\
\hline\hline
\textbf{Object Detection} &
\cite{chen2019detnas,ghiasi2019fpn,tan2020efficientdet} &
\cite{xu2019auto,yao2020sm,fang2020fast,guo2020hit,jiang2020elixirnet,xiong2020mobiledets,jiang2020sp,chen2020mnasfpn} \\
\hline
\textbf{Semantic Segmentation} &
\cite{chen2018searching,liu2019auto} &
\cite{nekrasov2019fast,nekrasov2020architecture,nekrasov2020template,lin2020graph,wong2019segnas3d,fang2020fast,zhang2020dcnas} \\
\hline
\textbf{Generative Models} &
\cite{gong2019autogan} &
\cite{gao2020adversarialnas,li2020gan,fu2020autogan,tian2020alphagan,zhou2020searching,tian2020off} \\
\hline
\textbf{Medical Image Analysis} &
\cite{weng2019unet} &
\cite{mortazi2018automatically,baldeon2020adaresu,zhu2019v,bae2019resource,zhang2019neural,kim2019scalable,dong2019neural,yu2020c2fnas,guo2020organ,yan2020ms} \\
\hline
\textbf{Adversarial Learning} &
-- &
\cite{vargas2019evolving,dong2019neural,guo2020meets} \\
\hline
\textbf{Low-level Vision} &
-- &
\cite{van2019evolutionary,chu2019fast,chu2019multi,liu2019deep,song2020efficient,guo2020hierarchical,mozejko2020superkernel,li2020all,zhang2020memory,lee2020journey} \\
\hline
\textbf{Video Processing} &
-- &
\cite{piergiovanni2019evolving,ryoo2020assemblenet,peng2019video,qiu2019scheduled} \\
\hline
\textbf{Multi-task Learning} &
-- &
\cite{perez2019mfas,yu2020software,liu2019training,gao2020mtl} \\
\hline
\textbf{Language Modeling} &
\cite{so2019evolved} &
\cite{chen2018exploring,wong2018transfer,veniat2019stochastic,maziarz2018evolutionary,mazzawi2019improving,jiang2019improved,wang2020textnas,chen2020adabert,li2020learning2} \\
\hline
\textbf{Speech Recognition} &
-- &
\cite{baruwa2019leveraging,li2019neural,ding2020autospeech} \\
\hline
\textbf{Graph Networks} &
-- &
\cite{gao2019graphnas,zhou2019auto,zhao2020probabilistic} \\
\hline
\textbf{Recommendation} &
-- &
\cite{li2020autost,cheng2020differentiable,zhao2020amer,song2020towards} \\
\hline
\textbf{Miscellaneous} &
\cite{miikkulainen2019evolving} (\textbf{captioning}) &
\cite{huang2018gnas} (\textbf{attribute}), \cite{quan2019auto,zhang2019person} (\textbf{person re-id}), \cite{li2019architecture} (\textbf{inpainting}), \cite{yang2019pose} (\textbf{pose}), \cite{zhang2019identify} (\textbf{reconstruction}), \cite{zhang2019neural2} (\textbf{knowledge graph}), \cite{an2020ultrafast} (\textbf{style transfer}), \cite{bianco2020neural} (\textbf{saliency}), \cite{zhang2019preventing} (\textbf{privacy}), \cite{ho2020neural} (\textbf{deep image prior}), \cite{hu2020count} (\textbf{counting}), \cite{zhang2020efficient} (\textbf{text recognition}), \cite{kokiopoulou2019fast} (\textbf{discussion}) \\
\hline
\end{tabular}
\caption{Summary of search spaces. Note that (i) there are two categorization methods, \textit{i.e.}, by space configuration (NAS-RL, NASNet, MobileNet, DARTS, open, others) and by non-classification task (detection, segmentation, \textit{etc.}); (ii) one work may appear in multiple rows because it deals with more than one space or tasks. We use both the total number of Google scholar citations and the number per time unit, according to the statistics on the July 31st, 2020, to judge if this work is an `important work' (around 10\% top-cited papers are considered important).}
\label{tab:search_spaces}
\end{table*}

\subsubsection{The \textbf{MobileNet} Search Space}
\label{formulation:space:MobileNet}

An alternative way~\cite{cai2019proxylessnas,wu2019fbnet} of simplifying the NAS-RL search space is to borrow the design of MobileNet. The macro architecture is set to be \textbf{chain-styled}, \textit{i.e.}, each layer receives input from its direct precursor. The micro architecture is also constrained to be one of the MobileNet (MB) cells~\cite{howard2017mobilenets}. Changeable options include the type of separable-convolution~\cite{chollet2017xception} (the regular convolution can be replaced by de-convolution), the type of skip-layer connection (can be either identity or some kinds of pooling), the number of channels, the kernel size, the expansion ratio (how the intermediate layer of the MB cell magnitudes the number of basic number of channels). Beyond these, the number of layers within each cell (or referred to as a block) can also be searched~\cite{tan2019mnasnet,fang2020densely}. So, the number of architectures in each cell is basically $C^1+C^2+\ldots+C^L$ where $C$ is the number of combinations of each layer and $L$ is the maximal number of layers in a cell. The above number is further powered up by $N$, the number of cells in the final architecture.

\noindent\textbf{Related search spaces.}\quad MNASNet~\cite{tan2019mnasnet} was the predecessor of the MobileNet search space, where there are more searchable options including the number of cells in each stage, \textit{i.e.}, $L$ is relatively large. After that, researchers soon noticed that (i) there are duplicates in using different numbers of layers in neighboring cells and (ii) preserving a small number of MB cells is sufficient to achieve good performance. So, a few follow-up methods~\cite{tan2019efficientnet,howard2019searching,chu2019fairnas} have reduced $L$ into $1$ and reduced $C$ into $6$ (\textit{i.e.}, MB cells with a $3\times3$, $5\times5$, or $7\times7$ convolution, and an expansion ratio of $3$ or $6$). The MobileNet space is also friendly to augmentation, including introducing mixed convolution~\cite{tan2019mixconv,chen2019efficient,mei2020atomnas}, allowing dense connections between layers~\cite{fang2020densely,guo2020powering}, or adding attention blocks~\cite{li2020neural}.

\subsubsection{Other Search Spaces}

Besides the previous search spaces that were widely explored, there exist other interesting methods. For example, some early efforts have either applied the technique of network transformation~\cite{chen2016net2net} or morphism~\cite{wei2016network} to generate complex architectures from the basic one~\cite{miikkulainen2019evolving,cai2018efficient,liang2018evolutionary}, or made use of shared computation to combine exponentially many architectures with cheap computation~\cite{wang2016deeply,saxena2016convolutional,zhao2018deep}. Researchers have also designed other methods for architecture design such as repeating one-of-many choices~\cite{baker2017designing,cortes2017adanet} or defining a language for encoding~\cite{negrinho2017deeparchitect,negrinho2019towards}. However, in the current era, increasing the search space arbitrarily seems not the optimal choice, since the search strategy is not sufficiently robust and may produce weak results that are on par with random search~\cite{xie2019exploring}.

There are efforts in applying NAS methods to other tasks, for which the search spaces need to adjust accordingly. For \textbf{object detection}, there are efforts of transferring powerful backbones in classification~\cite{tan2020efficientdet}, searching for a new backbone~\cite{chen2019detnas,chen2020mnasfpn,du2020spinenet}, considering the way of fusing multi-stage features~\cite{ghiasi2019fpn,xu2019auto}, and integrating multiple factors into the search space~\cite{guo2020hit,yao2020sm}. Regarding \textbf{semantic segmentation} in which spatial resolution is more sensitive~\cite{sun2019deep}, the most interesting part seems to design of the down-sampling and up-sampling path~\cite{liu2019auto,wu2019sparsemask}, though searching for efficient blocks is also verified useful~\cite{weng2019unet,shaw2019squeezenas,lin2020graph}. NAS is also applied for \textbf{image generation}, in particular, designing network architectures for the generative adversarial networks (GAN)~\cite{gong2019autogan,wang2019agan,gao2020adversarialnas}. Recently, this method has been extended to conditional GAN where the generator varies among classes and thus the search space becomes exponentially larger~\cite{zhou2020searching}. Beyond the search spaces that are extendable (see Sections~\ref{formulation:space:NASRL} and~\ref{formulation:space:DARTS}), there were efforts that directly search for powerful architectures for \textbf{RNN design}~\cite{schrimpf2017flexible,rawal2018nodes,ororbia2019investigating,huang2019wenet} or \textbf{language modeling}~\cite{chen2018exploring,wong2018transfer,so2019evolved,wang2020textnas}.

\noindent\textbf{Table~\ref{tab:search_spaces} provides a summary of search spaces for classification and the spaces transferred to different tasks.}

\subsection{Search Strategy and Evaluation Method}
\label{formulation:search_evaluation}

We combine the review of search strategies and evaluation methods into one part, because these two factors are closely correlated, \textit{e.g.}, the individual heuristic search strategies often lead to an separate evaluation on each sampled architecture, while the weight-sharing search strategies often integrate architecture optimization into architecture search.

\subsubsection{Individual Heuristic Search Strategies}
\label{formulation:search_evaluation:individual}

\begin{figure}[!t]
\centering
\includegraphics[width=8cm]{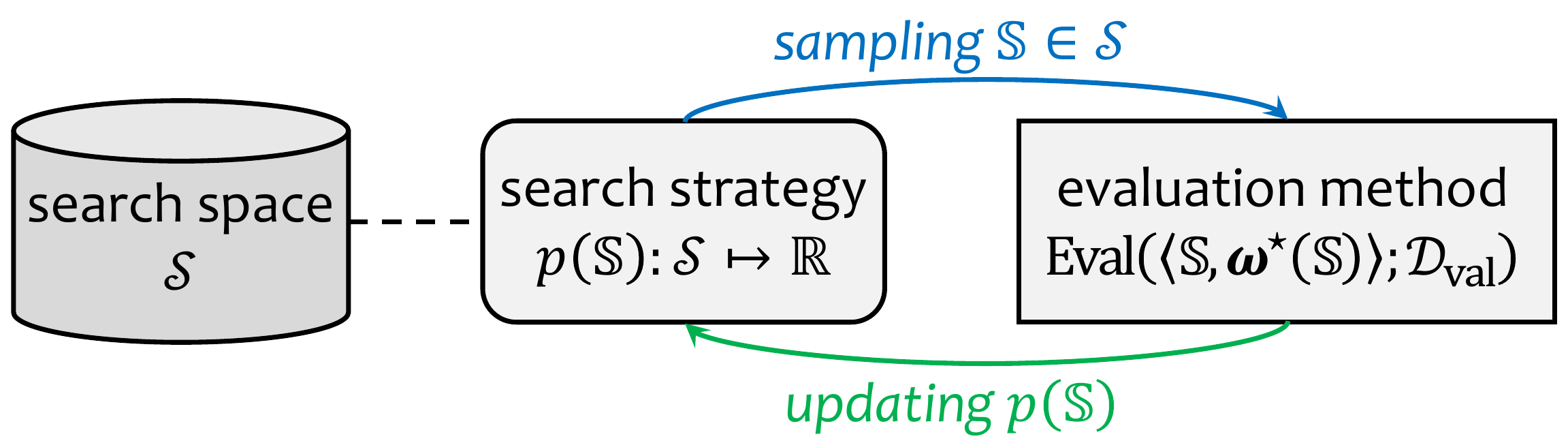}
\caption{The heuristic search method that maintains a probabilistic distribution on the search space. In each iteration, the algorithm samples an architecture following the distribution, performs evaluation individually, and updates the distribution accordingly. The design of this figure borrows the idea from~\cite{elsken2019neural}.}
\label{fig:search_evaluation}
\end{figure}

The most straightforward idea of architecture search is to enumerate all possible architectures from the search space and evaluate them individually. However, since the search space is often very large so that exhaustive search is intractable, researchers turned to heuristic search. The fundamental idea is to make use of the relationship between architectures, so that if an architecture reports good/bad performance, it is natural to tune up/down the probability of sampling its close neighbors. The algorithm starts with formulating a function, $p:\mathcal{S}\mapsto\mathbb{R}$, so that each architecture $\mathbb{S}$ in the search space corresponds to a probability density that it is sampled, and the function gets updated after the algorithm gets rewards from the evaluation result of the sampled architectures. This basic framework is illustrated in Figure~\ref{fig:search_evaluation}.

There are two popular choices of the heuristic algorithm, both of which maintain a latent form of the probabilistic distribution $p\!\left(\cdot\right)$. The first type, named the \textbf{genetic algorithm} or \textbf{evolutionary algorithm}, constructs a family of architectures during the search procedure~\cite{real2017large,xie2017genetic}, and allows each architecture (or named an individual) to generate other architectures sharing similar properties. Typical genetic operations include \textsf{crossover}, \textsf{mutation}, \textsf{selection} or \textsf{elimination}, \textit{etc}. The other type follows a different path, known as \textbf{reinforcement learning}~\cite{zoph2017neural,baker2017designing,zoph2018learning}, which constructs an architecture using a learnable policy, and updates the parameters in the policy according to the final performance. There have been efforts in combining the evolutionary algorithm and reinforcement learning for NAS~\cite{chen2019renas}, or comparing the efficiency between these two options~\cite{maziarz2018evolutionary} and/or against the random search baseline~\cite{real2019regularized}.

Regardless of the heuristic algorithm being used, such search methods suffer the problem of heavy computational overhead. This is mainly due to the need of re-training each sampled architecture from scratch. There are a few standard ways of alleviating the computational burden in the search procedure, including using a smaller number of channels, stacking fewer cells, or exploring on a smaller dataset. However, it can still take long for the algorithm to thoroughly traverse the search space and find a satisfying architecture, in particular when the training dataset is large.

\noindent\textbf{The random search baseline.}\quad
When the probability function does not change with the evaluation results, the individual heuristic algorithm degenerates to random search in which the probability of each architecture being sampled is equal and does not change with time. Note that random search can find any high-quality architecture given sufficient computational costs, but it is verified less efficient than heuristic search~\cite{real2019regularized} or weight-sharing search~\cite{bender2020can} methods. Random search is widely used as a baseline test of search efficiency, \textit{e.g.}, evaluating the difficulty of the search space and thus validating the high cost-performance ratio of the proposed approach~\cite{liu2019darts,li2019random,zela2020understanding,bi2020gold}.

\subsubsection{Weight-Sharing Heuristic Search Strategies}
\label{formulation:search_evaluation:weight_sharing}

A smart solution of fast search lies in sharing computation among different architectures, as many of them have similar properties, \textit{e.g.}, beyond a well-trained architecture, only one layer is added or one layer is made wider. In these scenarios, it is possible to copy the weights of the trained network with a small part being transformed~\cite{cai2018efficient}, so that the number of epochs required for optimizing the new network is significantly reduced.

Moving one step forward, an elegant solution of weight-sharing search is to construct the search space into a \textbf{super-network}~\cite{saxena2016convolutional,brock2017smash,pham2018efficient}, from which each architectures can be sampled as a \textbf{sub-architecture}, and different sub-architectures share corresponding modules. Consider a network that has a fixed topology with $L$ edges and each edge has $K$ options to be considered. The super-network contains $LK$ units to be optimized, but after the optimization, there are up to $K^L$ sub-architectures that can be sampled from the super-network. The search procedure is therefore largely accelerated because a large amount of computation has been shared among the sub-architectures.

However, it is worth emphasizing that sampling a sub-architecture with the weights directly copied from the super-network does not reflect its accurate performance on the validation set, because the weight-sharing training procedure was focused on optimizing the super-network as an entire but not the individual sub-architectures. This difficulty is named the optimization gap which we will formalize in Section~\ref{weight_sharing:gap}. To alleviate the gap, typical efforts include fine-tuning the sub-architectures for more accurate performance prediction~\cite{pham2018efficient}, facilitating the fairness during super-network training~\cite{chu2019fairnas}, and modeling the search space using graph-based methods~\cite{chen2020fitting}.

\noindent\textbf{Relationship to individual heuristic search.}\quad Note that, despite the super-network that organizes the search space in a more efficient manner, there still needs a standalone module to sample sub-architectures from the super-network. That is to say, either the evolutionary algorithm~\cite{guo2019single,chu2019fairnas} or the reinforcement learning algorithm~\cite{pham2018efficient,bender2018understanding} can be used. Due to the effectiveness of sub-architecture sampling, it is also possible to split the search space into subspaces and enumerate all architectures in each of them~\cite{chen2020fitting}.

\begin{table*}[!t]
\centering
\begin{tabular}{|L{4cm}||C{3cm}|C{9cm}|}
\hline
\textbf{Search Strategy} & \textbf{Important Work} & \textbf{Other Work} \\
\hline\hline
\textbf{Individual -- Reinforcement} &
\cite{zoph2017neural,baker2017designing,negrinho2017deeparchitect,zoph2018learning,zhong2018practical,liu2018progressive,tan2019mnasnet,ghiasi2019fpn,tan2019efficientnet,tan2020efficientdet} &
\cite{schrimpf2017flexible,dutta2018effective,dong2018dpp,hsu2018monas,zhou2018resource,mortazi2018automatically,perez2018efficient,zhong2020blockqnn,cheng2018searching,bashivan2019teacher,chen2018exploring,nekrasov2019fast,wong2018transfer,cheng2020instanas,guo2019irlas,jiang2019accuracy,perez2019mfas,gao2019graphnas,nekrasov2020architecture,nekrasov2020template,li2019architecture,gong2019autogan,zhou2019auto,balaprakash2019scalable,bae2019resource,zhang2019neural,baruwa2019leveraging,zhang2019preventing} \\
\hline
\textbf{Individual -- Evolutionary} &
\cite{real2017large,miikkulainen2019evolving,xie2017genetic,suganuma2017genetic,liu2018hierarchical,real2019regularized,stanley2019designing,so2019evolved} &
\cite{sun2018evolving,liang2018evolutionary,rawal2018nodes,prellberg2018lamarckian,lorenzo2018memetic,sun2019evolving,wang2018hybrid,lu2019nsga,piergiovanni2019evolving,maziarz2018evolutionary,van2019evolutionary,chu2019fast,chu2019multi,fang2019eat,ororbia2019investigating,liu2019deep,yu2020software,baldeon2020adaresu,ryoo2020assemblenet,saltori2019regularized,martens2019neural,irwin2019graph,shen2019searching,mazzawi2019improving,song2020efficient,zhang2019identify,an2020ultrafast,hassantabar2019steerage,bianco2020neural,ho2020neural,real2020automl,song2020towards} \\
\hline
\textbf{Individual -- Others} &
\cite{cortes2017adanet,chen2018searching,kandasamy2018neural,elsken2019efficient,cai2018path,xie2019exploring} &
\cite{smithson2016neural,wei2017modularized,wang2018evolving,sun2018particle,kamath2018neural,huang2018gnas,jin2018efficient,rohekar2018constructing,chen2019renas,sun2018automatically,fielding2018evolving,geifman2019deep,wang2019alphax,xiong2019resource,huang2019wenet,junior2019particle,fedorov2019sparse,radosavovic2019network,ma2019deep,wang2019particle,zhang2020autoshrink,jiang2020efficient} \\
\hline\hline
\textbf{Weight-Sharing Heuristic} &
\cite{cai2018efficient,brock2017smash,elsken2017simple,pham2018efficient,bender2018understanding,guo2019single,chu2019fairnas,cai2020once} &
\cite{wistuba2017finding,wistuba2018deep,zhang2019graph,zhang2018you,veniat2019stochastic,laube2019shufflenasnets,dikov2019bayesian,savarese2019learning,chen2019detnas,li2019partial,yu2019network,macko2019improving,wang2019hybrid,wong2019attonets,wistuba2019inductive,adam2019understanding,zhu2019eena,jiang2019neural,fang2020densely,wang2019sample,pasunuru2019continual,cheng2019swiftnet,vargas2019evolving,zhou2020posterior,tan2019mixconv,shaw2019squeezenas,chu2020moga,sun2019automatic,yu2020bignas,liu2019training,huang2019neural,li2020improving,zhao2019efficient,cheng2019msnet,li2020block,yao2020sm,kang2020towards,guo2020meets,mei2020atomnas,yu2020c2fnas,wang2020textnas,li2019neural,chen2019efficient,zhou2020econas,chu2020mixpath,ding2020bnas,meng2020neural,li2020gan,yu2020train,dai2020data,huang2020ponas,shen2020bs,you2020greedynas,zhang2020efficient,li2020neural,chen2020fitting,mozejko2020superkernel,guo2020organ,hu2020angle,ottelander2020local,xiong2020mobiledets,guo2020powering,lu2020neural,phan2020binarizing,wang2020dc,xia2020hnas,bender2020can,jiang2020sp,zaidi2020neural,ardywibowo2020nads,li2020gp,liu2020memnas,lyna2020fast,trofimov2020multi,zhao2020amer,zhou2020searching,cheng2020nasgem,lee2020journey,lu2020nsganetv2,tian2020off} \\
\hline
\textbf{Weight-Sharing Differentiable} &
\cite{liu2019darts,luo2018neural,cai2019proxylessnas,wu2019fbnet,xie2019snas,liu2019auto,li2019random,dong2019searching,chen2019progressive,nayman2019xnas,xu2020pc,zela2020understanding,liang2019darts+} &
\cite{shin2018differentiable,wu2018mixed,casale2019probabilistic,kokiopoulou2019fast,quan2019auto,hundt2019sharpdarts,stamoulis2019single,weng2019unet,noy2020asap,chen2019automatic,weng2019automatic,han2019design,zhou2019bayesnas,dong2019network,akimoto2019adaptive,zheng2019multinomial,saikia2019autodispnet,hu2019efficient,yao2020efficient,chang2019data,nguyen2020constrained,zheng2019dynamic,zhu2019v,cho2019one,peng2019video,zhou2019epnas,zhang2019efficient,chu2019scarletnas,yan2019hm,yang2020cars,li2019stacnas,carlucci2019manas,lin2020graph,luo2019understanding,qiu2019scheduled,wong2019segnas3d,yang2019pose,xu2019auto,guo2019nat,dong2019one,kim2019scalable,bi2019stabilizing,dong2019neural,lee2020efficient,zhang2019person,chu2020fair,jiang2019improved,li2020sgas,chen2020binarized,dong2020automatic,zhang2019neural2,green2019rapdarts,hong2019edas,xie2019exploiting,gao2020adversarialnas,jin2019rc,vahdat2020unas,wang2019scalable,chen2020adabert,fang2020fast,xu2020latency,hu2020dsnas,singh2020learning,chen2020stabilizing,he2020milenas,guo2020hit,guo2020hierarchical,gao2020mtl,liu2020labels,niu2020disturbance,bulat2020bats,hu2020count,jiang2020elixirnet,zhang2020dcnas,zhang2020adwpnas,zhao2020probabilistic,wan2020fbnetv2,li2020geometry,ding2020autospeech,chu2020noisy,li2020learning2,yang2020hournas,zheng2020rethinking,zhuo2020cp,li2020autost,chaudhuri2020fine,chen2020drnas,cheng2020differentiable,dai2020fbnetv3,fu2020autogan,li2020all,tian2020alphagan,wang2020m,yu2020cyclic,zhang2020memory,zhao2020few,zhou2020theory,bi2020gold,guo2020breaking,hong2020dropnas,kaplan2020self,li2020neural3,tian2020discretization,wang2020mergenas,wang2020si,yan2020ms,zhang2020one} \\
\hline
\textbf{Predictor-based Search} &
\cite{baker2017accelerating,deng2017peephole,dai2019chamnet} &
\cite{istrate2019tapas,sun2019surrogate,liu2019memnet,white2019bananas,xu2019renas,shi2019multi,friede2019variational,wei2020npenas,ning2020generic,tang2020semi,li2020neural2,mellor2020neural,nguyen2020optimal,ru2020revisiting,yan2020does,chau2020brp,luo2020neural,white2020study} \\
\hline
\end{tabular}
\caption{Summary of search strategies. We categorize the individual heuristic search strategies into three parts, \textit{i.e.}, using reinforcement learning, using evolutionary algorithms, and using other methods (\textit{e.g.}, Bayesian optimization, \textit{etc.}). Note that the categorization is not absolute, since the boundary between different methods is not always clear, \textit{e.g.}, an individual search method may apply weight-sharing in a minor part, and `other' individual methods may apply the ideology of reinforcement learning or evolutionary algorithms. For the judgment of `important work', please see the caption of Table~\ref{tab:search_spaces}.}
\label{tab:search_strategies}
\end{table*}

\subsubsection{Weight-Sharing Differentiable Search Strategies}
\label{formulation:search_evaluation:differentiable}

An alternative way of weight-sharing search is to allow a relaxed formulation of the super-network~\cite{luo2018neural,shin2018differentiable,liu2019darts} so that it is differentiable to both the network weights (mainly the convolutional kernels and batch normalization~\cite{ioffe2015batch} coefficients) and architectural parameters (modeling the likelihood that the candidate edges and/or operators are preserved in the final architecture). The most intriguing property of this formulation is that the architectural parameters can be optimized together with the network weights, so that at the end of the search procedure, the optimal sub-architecture is directly produced without the need of performing an additional sampling procedure, therefore, the computational costs are largely reduced.

The most popular example of differentiable search is DARTS~\cite{liu2019darts}, where the goal is to find the optimal sub-architecture for the normal and reduction cells as defined in Section~\ref{formulation:space:DARTS}. At the start, each edge is individually initialized so that all candidate operators have the same weight (\textit{e.g.}, $1/C$ where $C$ is the number of candidate operators). Then, during the search procedure, these architectural parameters are updated in a continuous space, so that the stronger (more effective) operators gradually take the lead. At the end of the search procedure, each edge undergoes a so-called discretization process, during which the strongest operator is preserved (and assigned with the weight of $1.0$) while others are eliminated. The preserved sub-architecture (the architectural parameters are fixed) is fed into the re-training procedure for the optimal network weights.

\noindent\textbf{Relationship to weight-sharing heuristic search.}\quad First, we emphasize that weight-sharing differentiable search also works on the MobileNet search space~\cite{cai2019proxylessnas,wu2019fbnet}, though it is more commonly used to explore the DARTS space. Essentially, differentiable search serves as an alternative way of approximately evaluating exponentially many architectures at the same time. The evaluation procedure is embedded into the search procedure and thus not explicitly performed. Therefore, the issue of optimization gap also exists in this framework. In particular, due to the inevitable approximation error of gradient computation, the differentiable search procedure can sometimes run into a dramatic failure~\cite{bi2019stabilizing,liang2019darts+}, and researchers developed various methods to alleviate this problem including performing early termination~\cite{liang2019darts+,zela2020understanding} or using progressive optimization~\cite{chen2019progressive,li2020sgas} to reduce the discretization error~\cite{tian2020discretization}. We will summarize these methods from another perspective in Section~\ref{weight_sharing:solutions}.

\subsubsection{Predictor-Based Search Methods}
\label{formulation:search_evaluation:predictors}

Besides the aforementioned search methods, there are also efforts in exhaustively enumerating all architectures in the search space (often relatively smaller) and then using them as the ground-truth to test the efficiency of different search strategies. The most popular benchmarks include NAS-Bench-101~\cite{ying2019bench} that contains $\sim423\mathrm{K}$ architectures evaluated on CIFAR10 and NAS-Bench-201~\cite{dong2020bench} that contains $15\rm{,}526$ architectures evaluated on CIFAR10, CIFAR100, and ImageNet-16-120. We also noticed an extension of the similar idea to the NLP scope~\cite{klyuchnikov2020bench}.

These benchmarks offer the opportunities of training predictors, \textit{i.e.}, the input is an encoded neural architecture and the output is the network accuracy~\cite{deng2017peephole}. There have been efforts of exploring the internal relationship between architectures (an effective way is to design a good encoding method for the architectures~\cite{friede2019variational,ning2020generic,white2020study}) for better prediction performance, in particular under a limited number of sampled architectures~\cite{shi2019multi,xu2020renas}. The trained predictors can inspire algorithms beyond the toy search spaces~\cite{chen2020fitting,ning2020generic}. We will discuss more about this issue in Section~\ref{weight_sharing:solutions:learning}.

\noindent\textbf{Table~\ref{tab:search_strategies} provides a summary of the frequently used search strategies, where we find that weight-sharing search is the current mainstream and future direction.}

\subsection{Summary}
\label{formulation:summary}

There is a close relationship between the search space and the search strategy together with evaluation method. When the search space is small and `tidy' (\textit{i.e.}, similar architectures are closely correlated in performance), individual heuristic search methods often work very well. However, when the search space becomes larger and more flexible, individual search methods often require heavy computational burden that is not affordable to most researchers and developers. Thus, since researchers believe that larger search spaces are the future trend of NAS, it has become an important topic to design efficient and stabilized weight-sharing search strategies. In the next part, we will owe the instability of weight-sharing architecture search to the optimization gap and discuss the solution to this problem.

As a side note, there have been research~\cite{zhou2019searching,radosavovic2020designing} on designing the search space for architecture search or design. From the perspective of NAS, this is similar to the idea that gradually removes the candidate connections/operators during the search procedure~\cite{chen2019progressive,li2020sgas,bi2020gold}, or undergoes a two-stage NAS algorithm that the network topology and operators are determined separately~\cite{bi2019stabilizing,hu2019efficient,yu2020c2fnas}. We carefully put forward the opinion that these multi-stage algorithms reflect the instability and immaturity of NAS algorithms, and we advocate for an algorithm that can explore a very large search space in an end-to-end manner.

In evaluating the quality of the search space and/or algorithms, choosing a good benchmark as well as a set of proper hyper-parameters is very important for NAS research. We suggest the readers to refer to~\cite{lindauer2019best} for more practices of research on NAS. Though, it was believed that evaluating the performance of NAS methods is often hard~\cite{sciuto2019evaluating,yang2020evaluation,yu2020evaluating}. Different settings beyond supervised learning have been investigated in NAS, including like semi-supervised learning~\cite{luo2020semi}, self-supervised learning~\cite{kaplan2020self}, unsupervised learning~\cite{liu2020labels,yan2020does}, incremental learning~\cite{huang2019neural,gao2020efficient}, federated learning~\cite{he2020fednas,xu2020federated}, \textit{etc.}, showing the promising transferability of NAS methods. Last but not least, there are several toolkits for AutoML~\cite{thornton2013auto,jin2019auto,erickson2020autogluon,kandasamy2020tuning,zimmer2020auto} that can facilitate the reproducibility of NAS methods.

\section{Towards Stabilized Weight-Sharing NAS}
\label{weight_sharing}

In this section, we delve deep into the weight-sharing NAS algorithms and try to answer the key question: why do these methods suffer search instability? Here, by \textbf{instability} we mean that the same algorithm can produce quite different results, in terms of either the searched architecture or the test accuracy, when it is executed for several times~\cite{li2019random}. For this purpose, we start with formulating the weight-sharing NAS methods in Section~\ref{weight_sharing:formulation} and formalizing the optimization gap in Section~\ref{weight_sharing:gap}. Then, in Section~\ref{weight_sharing:solutions}, we summarize the existing solutions to alleviate the optimization gap, followed by a summary of unsolved issues in Section~\ref{weight_sharing:unsolved}.

\subsection{An Alternative Formulation of Weight-Sharing NAS}
\label{weight_sharing:formulation}

Weight-sharing NAS methods start with defining a super-network that contains all learnable parameters in the search space. This involves using a vectorized form, $\boldsymbol{\alpha}$, to encode each architecture ${\mathbb{S}}\in{\mathcal{S}}$, so that the super-network is parameterized using both $\boldsymbol{\alpha}$ and $\boldsymbol{\omega}$. We follow the convention~\cite{liu2019darts} to rewrite the super-network as ${\mathbf{y}}={\mathbf{F}\!\left(\mathbf{x};\boldsymbol{\alpha},\boldsymbol{\omega}\right)}$. Throughout the remaining part of this paper, we refer to $\boldsymbol{\alpha}$ and $\boldsymbol{\omega}$ as the architectural parameters and network weights, respectively. Though $\boldsymbol{\alpha}$ takes discrete values, a common practice is to slack the search space and correspond each element of $\boldsymbol{\alpha}$ to the probability that an operator appears in the final sub-architecture. Hence, finding the optimal sub-architecture $\mathbb{S}^\star$ 
comes down to finding the optimal architectural parameters $\boldsymbol{\alpha}$ and performing discretization (mapping the possibly continuous $\boldsymbol{\alpha}$ to a discrete form that corresponds to a valid $\mathbb{S}$), and thus Eqn~\eqref{eqn:goal_trainval} becomes:
\begin{eqnarray}
\label{eqn:goal_slacked}
{\mathbb{S}^\star}={\mathrm{disc}\!\left(\boldsymbol{\alpha}^\star\right)}, & {\boldsymbol{\alpha}^\star}={\arg\min_{\boldsymbol{\alpha}}\mathcal{L}\!\left(\boldsymbol{\omega}^\star\!\left(\boldsymbol{\alpha}\right),\boldsymbol{\alpha};\mathcal{D}_\mathrm{val}\right)} \\
\nonumber
\mathrm{s.t.} & {\boldsymbol{\omega}^\star\!\left(\boldsymbol{\alpha}\right)}={\arg\min_{\boldsymbol{\omega}}\mathcal{L}\!\left(\boldsymbol{\omega},\boldsymbol{\alpha};\mathcal{D}_\mathrm{train}\right)},
\end{eqnarray}
where ${\mathcal{L}\!\left(\boldsymbol{\omega},\boldsymbol{\alpha};\mathcal{D}_\cdot\right)}={\mathbb{E}_{\left(\mathbf{x},\mathbf{y}^\star\right)\in\mathcal{D}_\cdot}\!\left[\mathrm{CE}\!\left(\mathbf{F}\!\left(\mathbf{x};\boldsymbol{\alpha};\boldsymbol{\omega}\right),\mathbf{y}^\star\right)\right]}$ are the expected cross-entropy loss (computed in any training subset). We expect ${\mathrm{Eval}\!\left(\left\langle\mathbb{S},\boldsymbol{\omega}\!\left(\mathbb{S}\right)\right\rangle\middle\vert\mathcal{D}_\cdot\right)}={1-\mathcal{L}\!\left(\boldsymbol{\omega},\boldsymbol{\alpha};\mathcal{D}_\cdot\right)}$ when $\boldsymbol{\alpha}$ corresponds to $\mathbb{S}$, but $\mathcal{L}\!\left(\boldsymbol{\omega},\boldsymbol{\alpha};\mathcal{D}_\cdot\right)$ is more flexible in dealing with the continuous form of $\boldsymbol{\alpha}$. Both $\mathcal{D}_\mathrm{train}$ and $\mathcal{D}_\mathrm{val}$ are subsets of the training data but they are not necessarily different, \textit{e.g.}, in one-level optimization methods~\cite{li2019stacnas,bi2020gold}, both $\mathcal{D}_\mathrm{train}$ and $\mathcal{D}_\mathrm{val}$ are the entire training set. The discretization function, $\mathrm{disc}\!\left(\cdot\right)$, does not take an essential operation when the search space is not slacked, \textit{e.g.}, in the weight-sharing heuristic search, $\boldsymbol{\alpha}$ does not take continuous values and $\mathbb{S}^\star$ is directly derived from $\boldsymbol{\alpha}^\star$.

Note that Eqn~\eqref{eqn:goal_slacked} can be solved using non-weight-sharing methods where $\boldsymbol{\omega}\!\left(\boldsymbol{\alpha}\right)$ have individual weights for different $\boldsymbol{\alpha}$. For weight-sharing methods, the key to optimize Eqn~\eqref{eqn:goal_slacked} is to share $\boldsymbol{\omega}\!\left(\boldsymbol{\alpha}\right)$ over different $\boldsymbol{\alpha}$ and perform the optimization with respect to $\boldsymbol{\omega}$ and $\boldsymbol{\alpha}$ over all sub-architectures. In other words, this is to achieve an amortized optimum over all sub-architectures and $\boldsymbol{\alpha}$ plays the role of the amortizing distribution. With this formulation, we provide an alternative perspective in discriminating weight-sharing heuristic and differentiable search. The \textbf{heuristic search} methods often keep all elements of $\boldsymbol{\alpha}$ identical, \textit{i.e.}, during training the super-network, the distribution $\mathfrak{P}$ remains uniform over the entire search space. A standalone heuristic search procedure is required to find the best sub-architecture. On the contrary, the \textbf{differentiable search} methods slack the search space and update $\boldsymbol{\alpha}$ and $\boldsymbol{\omega}$ simultaneously, assuming that $\boldsymbol{\alpha}$ gradually approaches the one-hot distribution corresponding to the best sub-architecture. So, after the super-network is well-trained, the best sub-architecture is often determined using a greedy algorithm on $\boldsymbol{\alpha}$ while $\boldsymbol{\omega}$ is often directly discarded.

\subsection{The Devil Lies in the Optimization Gap}
\label{weight_sharing:gap}

\begin{figure*}
\centering
\includegraphics[width=16cm]{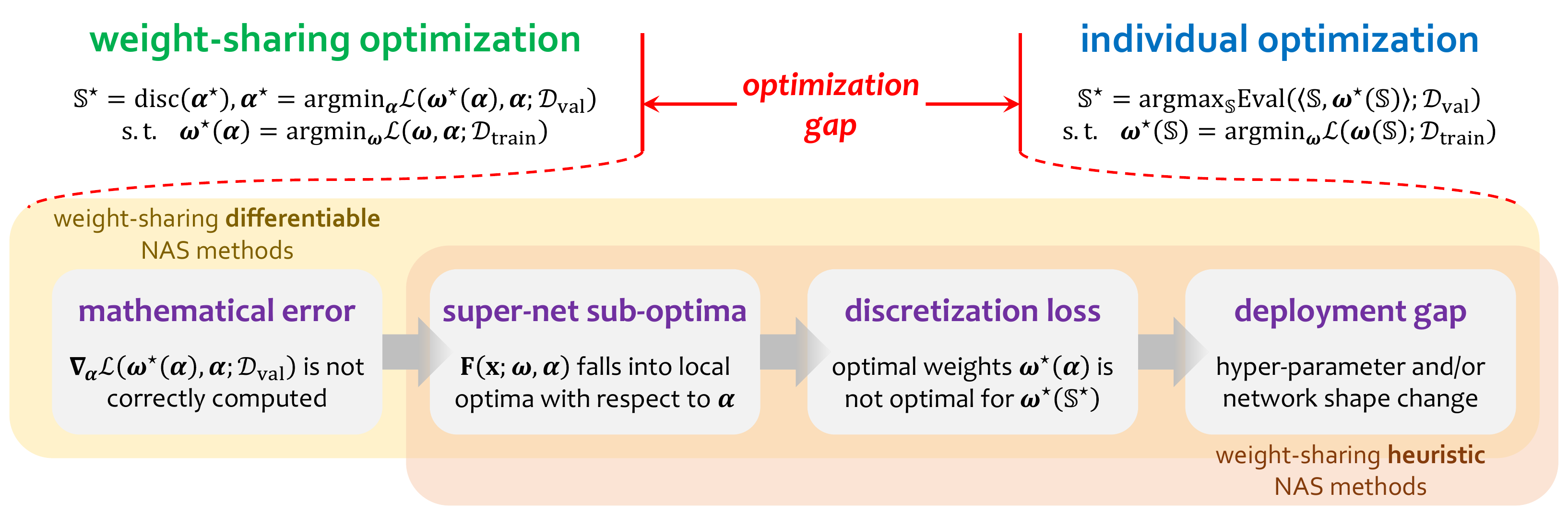}
\caption{The optimization gap between the goals of weight-sharing optimization and individual optimization, where the latter is accurate while the former is approximated. We summarize four major reasons to compose of the optimization gap, sorted by their appearance in the NAS flowchart and corresponding to the contents in Sections~\ref{weight_sharing:solutions:mathematics}--\ref{weight_sharing:solutions:deployment}, respectively. Note that weight-sharing heuristic methods may not involve the mathematical error because $\nabla_{\boldsymbol{\alpha}}\mathcal{L}\!\left(\boldsymbol{\omega}^\star\!\left(\boldsymbol{\alpha}\right),\boldsymbol{\alpha};\mathcal{D}_\mathrm{val}\right)$ is often not computed.}
\label{fig:optimization_gap}
\end{figure*}

Both types of weight-sharing search algorithms suffer the instability issue, in particular, the search results are sensitive to random initialization and/or hyper-parameters. We owe this problem to the \textbf{optimization gap}, \textit{i.e.}, a well-optimized super-network does not always produce high-quality sub-architectures. Mathematically, it implies that optimizing Eqn~\eqref{eqn:goal_slacked} yields an $\boldsymbol{\alpha}^\star$ and the corresponding ${\mathbb{S}^\star}={\mathrm{disc}\!\left(\boldsymbol{\alpha}^\star\right)}$, but there exists another architecture $\mathbb{S}'$ that achieves far higher accuracy. Below, we show examples of how the optimization gap appears as different phenomena according to the type of weight-sharing NAS algorithms.

For \textbf{weight-sharing heuristic search}, the optimization gap can be quantified by sampling a set of sub-architectures, $\left\{\mathbb{S}_1,\mathbb{S}_2,\ldots,\mathbb{S}_M\right\}$, and observing the true accuracy (by training each of them from scratch) and estimated accuracy (predicted by the super-network). Unfortunately, the relationship is often quite weak, \textit{e.g.}, the Kendall's $\tau$-coefficient between the rank lists according to the true and estimated accuracy is close to zero~\cite{zhang2020deeper,chen2020fitting} when the candidates have similar complexity. This is partly due to the strategy that optimizes part of the super-network in each iteration~\cite{cai2019proxylessnas,guo2019single,chu2019fairnas}, so that whether a sub-architecture performs well in the final evaluation depends on the frequency that it appears, in particular the final iterations of the training procedure. This is mostly random and uncontrollable.

On the other hand, for \textbf{weight-sharing differentiable search}, the optimization gap can be quantified by observing the accuracy of the optimal sub-architecture without re-training the network weights, $\boldsymbol{\omega}$. It is shown in~\cite{tian2020discretization} that even when the super-network achieves around $90\%$ accuracy (in the CIFAR10 training set), the pruned sub-architecture, without parameter re-training, often reports less than $20\%$ accuracy in the same dataset, which is dramatically low. This implies that the super-network is trained without `knowing' that a discretization stage will take effect to produce the final sub-architecture.

Summarizing the above, the optimization gap likely adds a non-observable noise, $\epsilon\!\left(\boldsymbol{\alpha}\right)$, to the true accuracy of ${\mathrm{Eval}\!\left(\boldsymbol{\alpha}\right)}\doteq{1-\mathcal{L}\!\left(\boldsymbol{\omega}^\star\!\left(\boldsymbol{\alpha}\right),\boldsymbol{\alpha};\mathcal{D}_\mathrm{val}\right)}$. Sometimes, for two candidates, say $\boldsymbol{\alpha}_1$ and $\boldsymbol{\alpha}_2$, the noise impacts even heavier, \textit{i.e.}, ${\left|\epsilon\!\left(\boldsymbol{\alpha}_1\right)-\epsilon\!\left(\boldsymbol{\alpha}_2\right)\right|}>{\left|\mathrm{Eval}\!\left(\boldsymbol{\alpha}_1\right)-\mathrm{Eval}\!\left(\boldsymbol{\alpha}_2\right)\right|}$. Due to the randomness of $\epsilon\!\left(\boldsymbol{\alpha}\right)$, the judgment becomes unreliable. This can largely downgrade the search results, or cause the results sensitive to initialization or hyper-parameters.

\subsection{Shrinking the Optimization Gap}
\label{weight_sharing:solutions}

Essentially, the optimization gap is the consequence of improving search speed with the price of lower evaluation accuracy. Shrinking the optimization gap forms the core challenge of weight-sharing NAS, for which researchers have proposed various solutions. Before going into details, we point out that the optimization gap is inevitable because the network weights are shared by exponentially many sub-architectures and the optimization procedure focuses on minimizing the amortized loss over the entire search space, not a specific sub-architecture.

We first summarize the typical reasons that cause the optimization gap into four categories, shown in Figure~\ref{fig:optimization_gap}. \underline{First}, when the differentiable methods are used, the end-to-end training procedure may suffer dramatic errors in estimating the gradients with respect to $\boldsymbol{\alpha}$. \underline{Second}, the search procedure may cause the super-network fall into a local 
optimum, in particular the joint optimization of $\boldsymbol{\alpha}$ and $\boldsymbol{\omega}$ increases the complexity of the high-dimensional space. \underline{Third}, the discretization process that derives the optimal sub-architecture may a dramatic accuracy drop on the network training accuracy. \underline{Fourth}, there are some differences in the network shape or hyper-parameters between the search and evaluation phases, making the searched architecture difficult in deployment. Note that most weight-sharing NAS approaches may involve more than one of the above issues. In what follows, we review the solutions to shrink the optimization gap by considering these factors individually or applying a learning-based approach that deals with the entire system as a black-box.

\subsubsection{Fixing the Mathematical Error}
\label{weight_sharing:solutions:mathematics}

The optimization error mostly exists in the differentiable search method, in particular, DARTS~\cite{liu2019darts}, in which a bi-level optimization method is used for updating $\boldsymbol{\omega}$ and $\boldsymbol{\alpha}$ separately. The reason for bi-level optimization lies in the imbalance between $\boldsymbol{\alpha}$ and $\boldsymbol{\omega}$. Most often, the number of learnable parameters of $\boldsymbol{\alpha}$ is typically hundreds or thousands, but the that of $\boldsymbol{\omega}$ is often millions. So, minimizing Eqn~\eqref{eqn:goal_slacked} can easily bias towards optimizing $\boldsymbol{\omega}$ due to the higher dimension of the parameter space\footnote{A direct consequence of such bias is that $\boldsymbol{\alpha}$ is not well optimized which introduces random noise to NAS. In an extreme situation, the super-network can achieve satisfying accuracy by only optimizing $\boldsymbol{\omega}$ (\textit{i.e.}, $\boldsymbol{\alpha}$ remains the randomly initialized status), but this delivers zero information to NAS.}. To avoid such bias, DARTS~\cite{liu2019darts} proposed to evaluate the network accuracy with respect to $\boldsymbol{\alpha}$ and $\boldsymbol{\omega}$ separately (\textit{i.e.}, ${\mathcal{D}_\mathrm{train}\cup\mathcal{D}_\mathrm{val}}={\varnothing}$), which leads to the iterative optimization:
\begin{eqnarray}
\label{eqn:bilevel}
{\boldsymbol{\omega}_{t+1}} & \leftarrow & {\boldsymbol{\omega}_t-\eta_{\boldsymbol{\omega}}\cdot\nabla_{\boldsymbol{\omega}}\mathcal{L}\!\left(\boldsymbol{\omega}_t,\boldsymbol{\alpha}_t;\mathcal{D}_\mathrm{train}\right)},\\
{\boldsymbol{\alpha}_{t+1}} & \leftarrow & {\boldsymbol{\alpha}_t-\eta_{\boldsymbol{\alpha}}\cdot\nabla_{\boldsymbol{\alpha}}\mathcal{L}\!\left(\boldsymbol{\omega}_{t+1},\boldsymbol{\alpha}_t;\mathcal{D}_\mathrm{val}\right)},
\end{eqnarray}
where $t$ is the iteration index starting with $0$, and $\eta_{\boldsymbol{\alpha}}$ and $\eta_{\boldsymbol{\omega}}$ are learning rates. Bi-level optimization strategy avoids the bias, but the correctness of bi-level optimization depends on (i) $\boldsymbol{\omega}_{t+1}$ has arrived at the optimum, \textit{i.e.}, ${\boldsymbol{\omega}_{t+1}}={\boldsymbol{\omega}^\star\!\left(\boldsymbol{\alpha}_t\right)}$, which is very difficult to satisfy especially when the dimensionality of $\boldsymbol{\omega}$ is high; (ii) the gradient with respect to $\boldsymbol{\alpha}$ is accurately computed, but this is computationally intractable due to the requirement of computing $\mathbf{H}_{\boldsymbol{\omega}}^{-1}$, the inverse Hessian matrix of $\boldsymbol{\omega}$. The solution that DARTS~\cite{liu2019darts} used is to ignore the optimality of $\boldsymbol{\omega}_{t+1}$ and use the identity matrix to approximate $\mathbf{H}_{\boldsymbol{\omega}}^{-1}$, but, as pointed out in~\cite{bi2019stabilizing,zela2020understanding}, this leads to dramatic errors in gradient computation\footnote{It was proved~\cite{bi2019stabilizing} that, even in optimizing a toy super-network, the angle between the true and estimated gradient vectors of $\boldsymbol{\alpha}$ can be larger than $90^\circ$, \textit{i.e.}, the optimization is going towards an uncontrolled, wrong direction.} and forms the major reason for mode collapse, \textit{e.g.}, the searched sub-architecture are mostly occupied by the \textsf{skip-connect} operator and perform quite bad~\cite{bi2019stabilizing,liang2019darts+,zela2020understanding,zhou2020theory}.

A straightforward solution to alleviate the search instability is to perform early termination~\cite{liu2019darts,liang2019darts+,zela2020understanding,wang2019scalable} or further constrain the property of the sub-architecture, \textit{e.g.}, containing a fixed number of \textsf{skip-connect} operators~\cite{chen2019progressive,liang2019darts+,wang2019scalable}. But, not allowing the search procedure to converge brings the problem that the search result is sensitive to initialization, which reflects in the unsatisfying stability, \textit{e.g.}, most differentiable search methods can run into bad sub-architectures with a probability of $20\%$--$80\%$.

There are some work that tried to stabilize bi-level optimization from the mathematical perspective. Typical efforts include using a better approximation ($\mathbf{H}_{\boldsymbol{\omega}}$ to replace $\mathbf{H}_{\boldsymbol{\omega}}^{-1}$) to achieve an estimation error bound~\cite{bi2019stabilizing}, applying adversarial perturbation as regularization to smooth the search space~\cite{chen2020stabilizing}, or applying mixed-level optimization to bypass the dramatic approximation~\cite{he2020milenas}. While these methods do not fix the error completely, an alternative path of research focuses on replacing bi-level optimization with one-level optimization which is free of the mathematical burden but the optimization bias towards $\boldsymbol{\omega}$ needs to be addressed. For this purpose, researchers proposed to partition the operator set into parameterized and non-parameterized groups to facilitates the fair comparison between similar operators~\cite{li2019stacnas}. A more elegant solution is provided by~\cite{bi2020gold} which shows that simply adding regularization (\textit{e.g.}, data augmentation~\cite{cubuk2019autoaugment} or Dropout~\cite{srivastava2014dropout}) or using a larger dataset (\textit{e.g.}, ImageNet~\cite{deng2009imagenet}) in the search procedure is sufficient to balance $\boldsymbol{\alpha}$ and $\boldsymbol{\omega}$ and improve one-level optimization. Nevertheless, since the validation set no longer exists, one-level optimization may suffers the difficulty of judging the extent that the super-network fits the training set, which may incur a higher risk of over-fitting.

\subsubsection{Avoiding the Sub-Optima of Super-Network}
\label{weight_sharing:solutions:regularization}

Training a super-network is not easy, because the space determined by $\boldsymbol{\alpha}$ is complicated and its property changes significantly with $\boldsymbol{\omega}$. This can be understood as the competition among different operators, and thus whether an operator is competitive largely depends on how its weights (part of $\boldsymbol{\omega}$) has been optimized. That being said, the operators that are easy to optimize (\textit{e.g.}, non-parameterized ones like \textsf{skip-connect} and \textsf{pooling} and the ones with fewer learnable weights like \textsf{convolution} with small kernels) may dominate in the early stage, which makes it even more difficult for the powerful operators (\textit{e.g.}, \textsf{convolution} of large kernels) to catch up with them~\cite{chen2019progressive}. In other words, the search algorithm runs into a local minimum in the space of $\boldsymbol{\alpha}$.

One of the effective solutions is to warm up the search procedure~\cite{chen2019progressive,xu2020pc,nayman2019xnas,xu2020latency}, \textit{i.e.}, freezing the update of $\boldsymbol{\alpha}$ in the starting stage and allowing $\boldsymbol{\omega}$ (mostly the weights of all \textsf{convolution} operators) to be well initialized. In essence, this is to postpone the optimization of $\boldsymbol{\alpha}$. In the heuristic search methods, it behaves as performing fair sampling~\cite{guo2019single,chu2019fairnas} for super-network training followed by a standalone sampling procedure starting with a uniform distribution on $\mathcal{S}$. Besides, there are also other kinds of regularization including partitioning the operators into parameterized and non-parameterized groups to avoid unfair competition~\cite{li2019stacnas,jiang2019neural}, using operator-level Dropout to weaken the superiority of non-parameterized operators~\cite{chen2019progressive}, and dynamically adjusting the candidate operator set to allow some perturbation in $\mathcal{S}$~\cite{nayman2019xnas}.

Note that bi-level optimization also belongs to the larger category of regularization, which decomposes the joint optimization into the subspaces of $\boldsymbol{\alpha}$ and $\boldsymbol{\omega}$. This explains why regularization plays an important role in one-level optimization~\cite{bi2020gold}. Finally, note that regularization often incurs slower convergence speed, but fortunately, the search cost of weight-sharing methods is relatively low and the increased overhead is often acceptable.

\subsubsection{Reducing the Discretization Loss}
\label{weight_sharing:solutions:discretization}

After the super-network has been well optimized, discretization plays an important role in deriving the optimal sub-architecture from the super-network, \textit{i.e.}, ${\mathbb{S}^\star}={\mathrm{disc}\!\left(\boldsymbol{\alpha}^\star\right)}$ as in Eqn~\eqref{eqn:goal_slacked}. A straightforward solution is to directly eliminate the weak operators from the super-network, but it has been shown that discretization can dramatically downgrade the training accuracy of the super-network (without re-training), especially for the differentiable search methods~\cite{chu2020fair,tian2020discretization}. There are mainly three solutions. \underline{First}, converting $\boldsymbol{\alpha}$ from continuous to discrete so that the search procedure is always optimizing the discretized sub-architectures. This strategy is used in both heuristic~\cite{guo2019single,chu2019fairnas} and differentiable~\cite{xie2019snas,casale2019probabilistic,chang2019data} search methods, and it can be interpreted as a kind of super-network regularization. \underline{Second}, pushing the values of $\boldsymbol{\alpha}$ towards either ends so that the eliminated elements have very low weights~\cite{noy2020asap,chu2020fair,tian2020discretization}. \underline{Third}, weakening one-time discretization as gradual network pruning~\cite{zheng2019dynamic,noy2020asap,li2020sgas,bi2020gold} so that the super-network is not pulled too far from the normal status ($\boldsymbol{\omega}$ is close to the optimum) and recovers after some training epochs (before the next pruning operation).

Note that, when the computational overhead is not considered, the safest way to reduce the discretization loss is to re-train all sub-architectures individually. Obviously, this strategy makes the algorithm degenerate to the heuristic search method with the distribution $\mathfrak{P}$ provided by the weight-sharing search procedure and the advantage in efficiency no longer exists. To make better use of the search procedure, a common practice is to fine-tune the top-ranked sub-architecture for a short period of time~\cite{cai2018efficient,white2019bananas}. The length of fine-tuning is an adjustable parameter and plays the role of a tradeoff between purely weight-sharing methods (when the length is zero) and purely individual methods (when the length goes to infinity).

\subsubsection{Bridging the Deployment Gap}
\label{weight_sharing:solutions:deployment}

The last step is to deploy the sub-architecture to the target scenario. In weight-sharing NAS, re-training is often required to achieve better performance, but there may be significant differences between the search and re-training settings. The major reasons are: (i) the search stage may need additional memory to save all network components and thus the network shape is shrunk; and (ii) the re-training stage may need to fit different constraints (\textit{e.g.}, the mobile setting~\cite{howard2017mobilenets} that requires the network FLOPs is just below $600\mathrm{M}$). A practical way is to adjust the network shape (\textit{e.g.}, the depth, width, and input resolution) and change the hyper-parameters (\textit{e.g.}, the number of epochs, the learning rate, whether some regularization methods have been applied, \textit{etc.}) accordingly. This causes a deployment gap. Conceptually, the algorithm finds the optimal sub-architecture under the setting used in the search procedure, but this does not imply its optimality under the hyper-parameters used in the re-training stage\footnote{For example, if the search stage contains much fewer cells, the searched cell architecture may have very deep connections which increases the difficulty of convergence in the re-training stage. Similarly, if the search stage is equipped with lighter regularization (\textit{e.g.}, Dropout), the searched architecture may contain fewer learnable parameters which downgrades its upper-bound in the re-training stage.}. Note that even with the same network shape, the difference in hyper-parameters can lead to unstable search results, while this factor is often concealed by other deployment gaps~\cite{bi2019stabilizing}.

The first revealed deployment gap is named the \textbf{depth gap}~\cite{chen2019progressive}, \textit{i.e.}, in the search space that contains repeatable cells, the searched sub-architecture can be extended in depth to achieve higher accuracy. However, it was shown that the optimal sub-architecture varies under different depths, and it proposed a progressive method to close the depth gap. Despite the progressively growing super-network, a more essential solution to the depth gap is to directly search in the target network shape. This is natural for most heuristic search methods~\cite{tan2019mnasnet,howard2019searching,chu2019fairnas,guo2019single} while the differentiable search methods can encounter the difficulty of training very deep super-networks -- this is mainly due to the mathematical error~\cite{bi2019stabilizing} elaborated in Section~\ref{weight_sharing:solutions:mathematics}. Going one step further, researchers have been investigating the relationship between the shallow super-network and the deep sub-architectures~\cite{meng2020neural,huang2020ponas,yu2020cyclic}, and a likely conclusion is that they can benefit from each other.

Another important deployment gap comes from transferring the searched architecture to other tasks, \textit{e.g.}, from image classification to object detection or semantic segmentation. Though many NAS methods have shown the ability of direct architecture transfer~\cite{zoph2018learning,chu2019fairnas,xu2020pc}, we argue that this is not the optimal solution since the optimal architecture may be task-specific. An example is that EfficientDet~\cite{tan2020efficientdet} is difficult to be trained yet does not show dominating detection accuracy, although it borrows the architecture and pre-trained weights from EfficientNet~\cite{tan2019efficientnet}, the state-of-the-art image classification architecture. To improve the transferability of the searched architectures, researchers have applied meta learning~\cite{robles2019learning,shaw2019meta,chen2020catch} (see Section~\ref{others:meta}) or performed network adaptation~\cite{fang2020fast}, but we point out that many aspects remain unexplored in this new direction.

\subsubsection{Learning-based Approaches}
\label{weight_sharing:solutions:learning}

To handle the complicated and coupled factors of the optimization gap, researchers have also investigated learning-based approaches that judge the quality of each sub-architecture based on both the super-network and other kinds of auxiliary information. We point out that this is closely related to the heuristic search methods that keep the distribution $\mathfrak{P}$ of the search space uniform. The difference lies in that the heuristic search methods often update a controller to latently assign a probability to each sub-architecture, but the learning-based methods often encode each sub-architecture into a vector and then use a model to predict the accuracy explicitly.

The performance of learning-based approaches largely depends on the way of representing the sub-architectures, as the encoding method determines the relationship between different sub-architectures~\cite{friede2019variational,ning2020generic,white2020study}. Sometimes, the auxiliary information does not come from a standalone predictor, but from a super-network, \textit{e.g.}, via knowledge distillation~\cite{cai2020once,trofimov2020multi}. Unlike formulating the entire space for the toy-sized NAS benchmarks~\cite{ying2019bench,dong2020bench}, in a large search space, the predictor is often embedded into the search procedure to guide the super-network optimization~\cite{white2019bananas,chen2020fitting,huang2020ponas}. Besides improving accuracy, the predictor can also filter out weak candidates earlier and thus accelerate the search procedure~\cite{zhou2019epnas,zheng2020rethinking}.

We emphasize that the learning-based approaches tend to consider NAS as a black-box predictor and use a small number of training samples to approximate the accuracy distribution on the search space. This is somewhat dangerous as the actual property of the distribution remains unclear. From a higher perspective, these approaches alleviate the optimization gap globally, but do not guarantee to capture the local property very well (\textit{i.e.}, the gap still exists).

\subsection{Unsolved Problems}
\label{weight_sharing:unsolved}

Despite the aforementioned efforts in shrinking the optimization gap, we notice that several important problems of weight-sharing NAS remain unsolved.

\underline{First}, weight-sharing NAS methods are built upon the assumption that the search space is relatively smooth. However, the actual property of the search space remains unclear. For example, does the search space contain exponentially many sub-architectures that are equally good? Given one of the best sub-architecture, $\mathbb{S}^\star$, is it possible that the neighborhood of $\mathbb{S}^\star$ is mostly filled with bad sub-architectures? These issues become increasingly important as the search space gets more complicated. But, due to the huge computational costs to accurately evaluate all sub-architectures (except for the toy search spaces~\cite{ying2019bench,dong2020bench,zela2020bench}), this problem remains mostly uncovered.

\underline{Second}, weight-sharing NAS methods facilitate different operators and connections to compete with each other, but it is unknown if the comparison criterion is made fair. For example, almost all search spaces involve the competition among \textsf{convolutions} with various kernel sizes (\textit{e.g.}, $3\times3$ vs. $5\times5$ or $7\times7$), and sometimes, network compression forces competition among different channel and/or bit widths. Although the `heavy' operators are intuitively stronger than the `light' operators, they often suffer slower convergence during the training procedure. That is to say, the search result is sensitive to the time point that comparison is made between these operators. This can cause considerable uncertainty especially in very large search spaces.

\underline{Third}, weight-sharing NAS methods are able to prune a large super-network into small sub-architectures, but may encounter difficulties in the reverse direction, \textit{i.e.}, deriving a large and/or deep sub-architecture from a relatively small super-network\footnote{Non-weight-sharing methods can achieve this goal~\cite{real2017large,real2020automl}, but the price is to re-train the enlarged network every time, bringing heavy computational overheads.}. This is necessary when higher recognition accuracy is desired, and existing approaches are mostly rule-based, including copying the repeatable cells (see Section~\ref{weight_sharing:solutions:deployment}) and applying simple magnification rules upon the base network. We expect weight-sharing NAS methods to have the ability of extending the search space when needed, which may require the assistance of dynamic routing networks (see Section~\ref{others:dynamic}).

\section{Related Research Directions}
\label{others}

We briefly review some related research directions to NAS, including (i) compressing the network towards a higher cost-performance ratio, (ii) designing dynamic routing network architectures so as to fit different input data, (iii) learning efficient learning strategies (meta learning), and (iv) looking for the optimal hyper-parameters for network training.

\subsection{Model Compression}
\label{others:compression}

As the deep networks become more and more complicated, deploying them to edge devices (\textit{e.g.}, mobile phones) requires a new direction named model compression~\cite{han2016deep,iandola2016squeezenet}. Two of the most popular techniques are network pruning~\cite{han2015learning,yang2017designing,sze2017efficient,huang2018data} that eliminates less useful channels or operators from a trained network, and network quantization~\cite{gong2014compressing,chen2015compressing,hubara2017quantized} that uses low-precision (\textit{e.g.}, $1$-bit~\cite{rastegari2016xnor}) computation to replace the frequently used \textsf{FP32} (\textit{i.e.}, $32$-bit floating point number) operations. Both methodologies have been verified to achieve a better tradeoff between accuracy and complexity.

Conceptually, model compression shares a similar formulation to NAS, \textit{i.e.}, the generalized formulation in Section~\ref{formulation:framework} directly applies with either a regularization term for model complexity or a hard constraint for the maximal resource. Therefore, NAS approaches are often easily transferred for model compression~\cite{ashok2018n2n,fedorov2019sparse}, including pruning~\cite{he2018amc,li2019partial,dong2019network,xiao2019autoprune,yu2019network,zhao2019efficient}, quantization~\cite{wu2018mixed,cai2019device,nascimento2020finding,yu2020search}, and joint optimization~\cite{chen2018joint,lu2020beyond,wang2020apq}. Sometimes, the searched configuration or connectivity mask can improve recognition accuracy~\cite{ahmed2018maskconnect}. Yet, the optimization gap also exists in these scenarios, and the methods introduced in Section~\ref{weight_sharing:solutions} are still useful, \textit{e.g.}, super-network regularization~\cite{xiao2019autoprune,wang2020pruning,li2020heterogeneity}, network fine-tuning~\cite{cai2019device}, re-training~\cite{gordon2018morphnet,chen2020network}, \textit{etc}. In addition, the unfairness issue is significant when the operators of different channel and/or bit widths exist~\cite{cai2020rethinking}. Nevertheless, 
it is believed that NAS will form an important module in the future direction named software-hardware co-design~\cite{cai2019automl,lu2019neural,abdelfattah2020best,gupta2020accelerator,jiang2020device,jiang2020hardware,jiang2020standing,lin2020mcunet}.

\subsection{Dynamic Routing Networks}
\label{others:dynamic}

Most neural architectures are static, \textit{i.e.}, not changing with input data. As a counterpart to save computation, researchers have investigated the option of adjusting the network architecture according to test data~\cite{liu2018dynamic}. This strategy, named dynamic routing, brings various benefits, including higher amortized efficiency by assigning heavy computation to easy samples and light computation to hard samples~\cite{huang2018multi,liu2018dynamic,wang2018skipnet} information for higher recognition accuracy~\cite{li2020learning}. There are also discussion on the training strategy for the dynamic routing networks~\cite{mcgill2017deciding,li2019improved}.

From a generalized perspective, a dynamic network can be considered as a variant of a static network. Note that even for a static network, the input data can cause part of neurons (after the activation function, \textit{e.g.}, ReLU~\cite{nair2010rectified}) to output a value of zero, that is, the connections to these neurons are eliminated. Therefore, the difference lies in that dynamic networks are switching off larger units (\textit{e.g.}, network layers or blocks~\cite{wu2018blockdrop}). This implies that NAS methods, with simple modification, can be applied for dynamic network design~\cite{yuan2019s2dnas,nekrasov2020architecture}. More importantly, dynamic routing networks may pave the way of weight-sharing NAS in an enlargeable search space~\cite{cortes2017adanet,zhang2019graph}, which we believe is a new research direction.

\subsection{Meta Learning}
\label{others:meta}

Meta learning~\cite{schmidhuber1987evolutionary,schaul2010metalearning} is sometimes referred to as `learning to learn', which aims to formulate the common practices in a learning procedure, so that the learned model can quickly adapt to new tasks and/or environments. It originates from the idea of mimicking the learning behavior of human beings~\cite{lake2017building}, yet eases the machine learning algorithms to deploy to various scenarios, \textit{e.g.}, the few-shot learning setting~\cite{vinyals2016matching,finn2017one,ren2018meta} that allows the model to learn new concepts with a small number of training samples. There are mainly three types of meta learners, namely, model-based~\cite{santoro2016meta,duan2016rl,mishra2018simple}, metric-based~\cite{vinyals2016matching,snell2017prototypical}, and optimization-based~\cite{li2016learning,ravi2017optimization,finn2017model} methods, differing from each other in the way of formulating the discriminative function\footnote{Borrowed from \textsf{http://metalearning-symposium.ml/files/vinyals.pdf}.}. For reviews of meta learning, please refer to~\cite{lemke2015metalearning,vanschoren2018meta}.

Meta learning has a longer history than NAS, and modern NAS was considered a special case of meta learning~\cite{baker2017accelerating,brock2017smash,elsken2017simple}, namely, learning to design a computational model that assists learning. In particular, meta learning plays an important role in the early age of NAS, when the NAS methods need to learn from a small number of sub-architectures that are sampled and individually evaluated. As weight-sharing NAS methods become more and more popular, meta learning can be used for various purposes, including (i) enhancing the few-shot learning ability of NAS methods~\cite{doveh2019metadapt,wang2020m,elsken2020meta}, (ii) allowing NAS methods to transfer to different tasks~\cite{robles2019learning,shaw2019meta,chen2020catch}, and (iii) providing complementary information to shrink the optimization gap~\cite{cheng2019swiftnet,dubatovka2019ranking}.

\subsection{Hyper-Parameter Optimization}
\label{others:HPO}

Hyper-parameter optimization (HPO) is another important direction of AutoML, with the aim of finding proper hyper-parameters to improve the quality of model training. This is a classic problem in machine learning~\cite{claesen2015hyperparameter}, but in the deep learning era, the optimization goal has been focused on the important settings that impact network training including the learning rate~\cite{andrychowicz2016learning,loshchilov2017sgdr,li2020exponential,loizou2020stochastic}, mini-batch size~\cite{mu2018parameter,yao2018hessian}, weight decay~\cite{loshchilov2019decoupled,zhang2019three}, momentum~\cite{sutskever2013importance,lancewicki2020automatic}, or a combination of them~\cite{mendoza2016towards,loshchilov2016cma,kandasamy2020tuning}. For more thorough reviews of HPO, please refer to~\cite{bergstra2011algorithms,feurer2019hyperparameter,yu2020hyper}. 

We point out that NAS is essentially a kind of HPO, \textit{i.e.}, the network architecture is encoded as a set of hyper-parameters. Indeed, the NAS approaches in the early years have mostly inherited the ideology of HPO methods including random search~\cite{bergstra2011algorithms,bergstra2012random}, heuristic search~\cite{snoek2012practical,bergstra2013making,maclaurin2015gradient,young2015optimizing,falkner2018bohb,li2017hyperband}, and meta learning~\cite{feurer2015initializing}. On the other hand, NAS approaches have been transplanted to HPO~\cite{bello2017neural}, and sometimes NAS and HPO are performed together~\cite{zela2018towards,klein2019tabular}. But, one of the important challenges that HPO algorithms have encountered is the much slower optimization speed compared to the weight-sharing NAS algorithms. This is because each step of HPO requires a complete training procedure, unlike NAS that updates the architectural parameters, $\boldsymbol{\alpha}$, in each step of the search procedure. Essentially, this is due to the missing of the \textbf{greedy optimization assumption}: during NAS training, it is always a good choice to minimize the loss function in a short period of time, but this does not hold for HPO training\footnote{A clear example lies in the learning rate. If the goal is to cause the loss function drop rapidly, the algorithm may choose to use a small learning rate at the start of training, but this strategy is not good for the global optimality.}. A direct consequence is the limited degree of freedom of the HPO search space. We expect the future research of HPO can alleviate this difficulty and thus achieve higher flexibility of optimizing hyper-parameters. The recent efforts to combine NAS and HPO~\cite{dong2020autohas} provided a promising direction.

In the field of AutoML, we notice an important subarea of HPO named automated data augmentation, which is currently one of the most effective and widely-applied HPO techniques. We introduce it separately in the next part.

\subsection{Automated Data Augmentation}
\label{others:AA}

Data augmentation is a technique that enlarges the training set using transformed copies of the existing training images. The assumption is that slightly perturbed (\textit{e.g.}, rotated, flipped, distorted, \textit{etc.}) samples share the same label as the original one. AutoAugment~\cite{cubuk2019autoaugment} first formulated data augmentation into an HPO problem in a large search space and applied reinforcement learning for finding the optimal strategy. It achieved good practice in many image classification datasets, but the search cost is very high (\textit{e.g.}, hundreds of GPU-days even for CIFAR10). There are a few follow-ups that tried to accelerate AutoAugment using approximated~\cite{lim2019fast,lin2019online,ho2019population} or weight-sharing~\cite{hataya2019faster} optimization, or simply used randomly generated strategies which were verified equally good~\cite{cubuk2020randaugment}. Without careful control, the aggressively augmented training samples may contaminate the training procedure. To weaken the impact of these samples and alleviate the empirical risk, researchers proposed to increase the training batch size~\cite{zhang2020adversarial}, compute individual batch statistics~\cite{xie2020adversarial}, or use knowledge distillation~\cite{wei2020circumventing} for case-specific training label refinement.

There is a survey~\cite{shorten2019survey} that reviews a wide range of data augmentation techniques for deep learning. Applying AutoML techniques for data augmentation was believed to have a high potential for future research.

\section{Conclusions and Future Prospects}
\label{future}

Neural architecture search (NAS) is an important topic of automated machine learning (AutoML) that aims to find replacement for the manually designed architectures. With three-year rapid development, NAS shows the potential of becoming a standard tool of deep learning algorithms applied to different AI problems. Currently, NAS has contributed significantly to some state-of-the-art computer vision algorithms, but NAS also suffers significant drawbacks that hinders its broader applications.

However, in the past months, the research of NAS seems to have slowed down, arguably due to both the limited search space and unstable search algorithms. These factors are highly correlated. The conservative design of the search spaces makes it difficult to evaluate and compare the performance of different search algorithms, on the other hand, the instability of the search algorithms obstacles the exploration in larger search spaces. Today, two most popular search spaces, the DARTS search space (Section~\ref{formulation:space:DARTS}) and the MobileNet search space (Section~\ref{formulation:space:MobileNet}) have been so thoroughly explored that some tricks (\textit{e.g.}, preserving a fixed number of \textsf{skip-connect} operators in each cell~\cite{chen2019progressive,liang2019darts+}) can produce highly competitive search results, or some inaccurate conclusions can be made~\cite{xie2019exploring,ottelander2020local}, yet these results do not show significant advantages over the random search methods with a reasonable amount of trials.

As the last but not least part of this survey, we share our opinions on the future directions of NAS. We believe that the future prospect of NAS lies in whether researchers can propose a new framework to systematically solve the aforementioned problems, in particular, designing more flexible search spaces and developing reliable search algorithms. These two factors facilitate each other, but in the current time point (July 31st, 2020), the former seems more urgent.

A good \textbf{search space} for NAS research is expected to satisfy four conditions. \underline{First}, the space is sufficiently large to include both chain-styled~\cite{simonyan2015very} and multi-path~\cite{szegedy2015going} networks, popular hand-designed modules (\textit{e.g.}, the residual block~\cite{he2016deep}, group convolution~\cite{xie2017aggregated}, depthwise separable convolution~\cite{chollet2017xception}, shuffle modules~\cite{ma2018shufflenet}, attention units~\cite{hu2018squeeze,zhang2020resnest}, activation functions~\cite{ramachandran2017searching,wang2019scalable}, \textit{etc.}), and the combination of them~\cite{howard2017mobilenets}. According to our rough estimation, the cardinality of such a search space is expected to exceed $10^{1000}$, but the largest search space to date (GOLD-NAS~\cite{bi2020gold}) only has a cardinality of $10^{110}$--$10^{170}$. \underline{Second}, the space should be adjustable, allowing the search algorithm to increase or decrease the model size without restarting the search procedure from scratch. \underline{Third}, the space can be applied to a wide range of vision problems without heavy modifications -- preferably, the searched architecture transfers to different tasks with merely light-weighted heads to be replaced (or searched), like HRNet~\cite{wang2020deep}. \underline{Fourth}, the best architecture in the space should achieve significantly better accuracy-complexity tradeoff than that found by random search (with reasonable computational costs) or manual design (including pure manual design and adding simple rules to the search procedure).

Based on the search space, a good \textbf{search algorithm} should, first of all, be able to explore the search space without getting lost, \textit{e.g.}, running into a set of sub-optimal architectures that are even worse than random search or manual design. We believe that various network training methods will be useful, including knowledge distillation~\cite{hinton2015distilling,such2019generative,gu2020search,liu2020search,trofimov2020multi}, automated data augmentation (see Section~\ref{others:AA}), dynamic routing networks (see Section~\ref{others:dynamic}), and transfer learning~\cite{lu2020neural,panda2020nastransfer}. In addition, the algorithm should allow different constraints (\textit{e.g.}, hardware factors like latency~\cite{wu2019fbnet,lyna2020fast,xu2020latency} and other factors~\cite{stamoulis2019single,fernandez2020searching}) to be embedded. We believe that such NAS algorithms, once developed, can largely facilitate the development of deep learning and even the next generation of AI computing.

\ifCLASSOPTIONcompsoc
  \section*{Acknowledgments}
\else
  \section*{Acknowledgment}
\fi

The authors would like to thank the colleagues in Huawei Noah's Ark Lab, in particular, Prof. Tong Zhang, Dr. Wei Zhang, Dr. Chunjing Xu, Jianzhong He, and Zewei Du, for instructive discussions. We would also like to thank the organizers of \textsf{AutoML.org} for providing an awesome list of AutoML papers\footnote{This survey is gratefully helped by the list of literature on \textsf{https://www.automl.org/automl/literature-on-neural-architecture-search/}.}.

\bibliographystyle{IEEEtran}
\bibliography{bare_jrnl_compsoc}

\end{document}